%% file: main.tex
\documentclass[10pt,twocolumn,letterpaper]{article}

\usepackage[pagenumbers]{cvpr} 

\usepackage[table]{xcolor} 
\usepackage[table]{xcolor}    
\usepackage{adjustbox}        
\usepackage{array}            
\usepackage{float}
\usepackage[misc]{ifsym} 
\usepackage{algorithm}
\usepackage{algorithmic}
\usepackage{multirow}
\usepackage{overpic}

\newcommand{\myparagraph}[1]{{\vspace{.3em} \noindent \bf #1}}
\usepackage{xspace}
\usepackage{pifont} %
\newcommand{\cmark}{\ding{51}\xspace}%
\newcommand{\xmarkg}{\textcolor{lightgray}{\ding{55}}\xspace}%
\newcommand{\videonum}{60K\xspace}
\newcommand{\begray}[1]{\textcolor{lightgray}{#1}}
\usepackage{makecell} 

\input{preamble}
\definecolor{cvprblue}{rgb}{0.21,0.49,0.74}
\usepackage[pagebackref,breaklinks,colorlinks,allcolors=cvprblue]{hyperref}

\def\paperID{} 
\def\confName{CVPR}
\def\confYear{2026}

\title{EffectErase: Joint Video Object Removal and Insertion\\ for High-Quality Effect Erasing}

\author{
Yang Fu \quad Yike Zheng \quad Ziyun Dai \quad
Henghui Ding~\!$^{\textrm{\Letter}}$
\vspace{.6mm}
\\
{\fontsize{11.6}{11}\selectfont Institute of Big Data, College of Computer Science and Artificial Intelligence, Fudan University, China}\\
\href{https://henghuiding.com/EffectErase/}{https://henghuiding.com/EffectErase/}
}

\begin{document}

\twocolumn[{
\maketitle
\input{fig/fig_teaser}
}]



\renewcommand{\thefootnote}{\fnsymbol{footnote}}
\footnotetext[0]{${\textrm{\Letter}}$ Corresponding author (henghui.ding@gmail.com).}

\input{sec/0_abstract}

\input{sec/1_intro}
\input{sec/2_related}

\input{sec/3_method}

\input{sec/4_exp}
\input{sec/5_con}

{
    \small
    \bibliographystyle{ieeenat_fullname}
    \bibliography{main}
}

{
\normalsize
\clearpage
\input{sec/X_suppl}
}

\end{document}

%% file: fig/fig_teaser.tex
\begin{center}
    \centering
    \captionsetup{type=figure}
    \vspace{-7mm}
    \includegraphics[width=\textwidth]{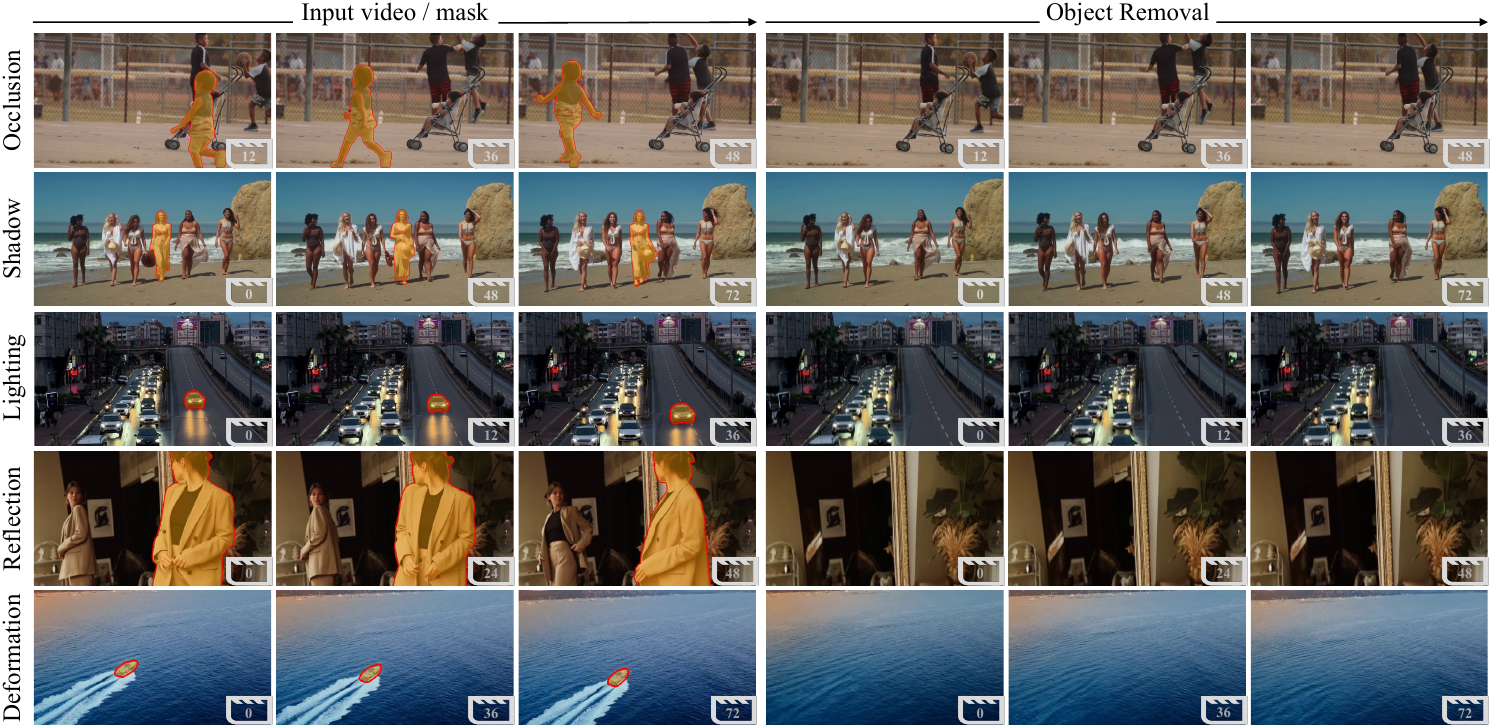}
    \vspace{-7mm}
    \captionof{figure}{
    \textbf{EffectErase} effectively removes target objects together with various object-induced effects in videos, such as occlusion, shadow, lighting, reflection, and deformation.
    }
    \label{fig:teaser}
\end{center}

%% file: sec/0_abstract.tex
\begin{abstract}

Video object removal aims to eliminate dynamic target objects and their visual effects, such as deformation, shadows, and reflections, while restoring seamless backgrounds.
Recent diffusion-based video inpainting and object removal methods can remove the objects but often struggle to erase these effects and to synthesize coherent backgrounds.
Beyond method limitations, progress is further hampered by the lack of a comprehensive dataset that systematically captures common object effects across varied environments for training and evaluation. To address this, we introduce \textbf{VOR} (\textbf{V}ideo \textbf{O}bject \textbf{R}emoval), a large-scale dataset that provides diverse paired videos, each consisting of one video where the target object is present with its effects and a counterpart where the object and effects are absent, with corresponding object masks.  
VOR contains \videonum high-quality video pairs from captured and synthetic sources, covers five effects types, and spans a wide range of object categories as well as complex, dynamic multi-object scenes.
Building on VOR, we propose \textbf{\textit{EffectErase}}, an effect-aware video object removal method that treats video object insertion as the inverse auxiliary task within a reciprocal learning scheme.
The model includes task-aware region guidance that focuses learning on affected areas and enables flexible task switching.
Then, an insertion-removal consistency objective that encourages complementary behaviors and shared localization of effect regions and structural cues.
Trained on VOR, EffectErase achieves superior performance in extensive experiments, delivering high-quality video object effect erasing across diverse scenarios.

\end{abstract}

%% file: sec/1_intro.tex
\vspace{-4mm}
\section{Introduction}
\label{sec:intro}

Video object removal has emerged as a key technique that enables users to erase unwanted dynamic content from videos while preserving realistic visual quality. It is widely used in film post-production and video editing.
Recent advances in generative models~\cite{brooks2024video,yang2024cogvideox,kong2024hunyuanvideo,wan2025wan}
have demonstrated remarkable progress in video generation and editing quality.~Leveraging the capabilities of large generative models, recent video object removal methods~\cite{li2025diffueraser,bian2025videopainter,zi2025minimaxremovertamingbadnoise,vace,miao2025rose} have shown promising performance across diverse scenarios.
However, as shown in~\cref{fig_limits_of_prev_inpainting}, these methods still struggle to achieve high-fidelity results when removing objects with complex visual effects such as reflections.

\input{fig/fig_limits_of_prev_inpainting}

This limitation can be attributed to the heavy reliance on the input mask for guidance in most video object removal methods~\cite{liu2021fuseformer,zhang2022flow,zhou2023propainter,li2025diffueraser,bian2025videopainter}, which often leads to overlooking the side effects that objects introduce into the scene.~To mitigate this issue, some methods, such as Minmax-Remover~\cite{zi2025minimaxremovertamingbadnoise}, implicitly trains the model to discover these effects, while ROSE~\cite{miao2025rose} explicitly predicts a difference mask for side effects and uses it as additional guidance.
However, they still lack explicit modeling of spatiotemporal correlations between objects and their effects, limiting their robustness in complex real-world scenes and preventing stable, precise localization of effect regions.

Beyond these methodological limitations, progress in this field is also limited by the lack of a large-scale and publicly available dataset that captures common object effects across various scenes. Recently, several image-based object removal datasets~\cite{Sagong_2022_BMVC,liu2024shadow,zhao2025objectclear} have been introduced to address the visual side effects caused by object, but they remain restricted to image-level, preventing video-based models from learning the temporal consistency required for handling moving objects. Constructing large-scale and diverse video datasets is more challenging, as the paired videos must maintain spatially consistent backgrounds and temporally coherent motion across frames.~SVOR~\cite{chang2019vornet} synthesizes video pairs by overlaying object masks from foreground videos in YouTube-VOS~\cite{xu2018youtube} onto background videos, but does not account for the visual side effects.
ROSE~\cite{miao2025rose} employs a 3D rendering engine to generate well-aligned synthetic video pairs, but it neglects object motion and relies solely on camera movement.
\textbf{New Dataset and Benchmark.}~To support research on effect-aware \textbf{V}ideo \textbf{O}bject \textbf{R}emoval in real-world scenarios, we construct \textit{\textbf{VOR}}, a large-scale hybrid dataset that combines camera-captured and 3D-synthesized videos featuring diverse foreground objects, background scenes, and object effects.
{For the camera captured data}, we use multiple tripod-mounted cameras to record paired videos across 293 scenes, broadly covering typical real-world use cases of video object removal. 
{For the synthesized data}, we construct over 150 diverse 3D scenes containing multiple dynamic objects, rendered by a 3D graphics engine.
To approximate real-world scenarios, we manually design realistic camera and object trajectories.
By combining the realism of camera-captured data with the diversity of synthesized content, VOR provides a high-quality, large-scale dataset comprising \videonum paired videos.
For a comprehensive evaluation of video object removal methods, we further introduce two benchmarks, \textit{VOR-Eval}, a curated set with ground truth, and \textit{VOR-Wild}, an in-the-wild set without ground truth covering a wide range of real-world videos.

\input{fig/fig_limits_of_prev_removal}

\input{tab/dataset_comp}

\textbf{EffectErase: Joint Removal–Insertion.} Motivated by the complementary relationship of video object removal and insertion, which operate on the same affected regions as shown in~\cref{fig_remove_joint}, 
we propose EffectErase, an effect-aware dual learning framework that jointly learns video object removal and insertion, treating insertion as an inverse auxiliary task to enhance removal quality.
EffectErase incorporates a Task-Aware Region Guidance (TARG) module and an Effect Consistency (EC) loss.
The TARG module builds spatiotemporal correlations between the target object and its side effects through a cross-attention mechanism, guiding the model to accurately identify the affected regions. In addition, a task token in this module enables flexible switching between the removal and insertion tasks.
EC encourages the two inverse tasks to share consistent effect regions and structural feature representations, enforcing cross-task consistency and strengthening effect-aware learning. Together, these components allow EffectErase to accurately localize and erase visual side effects across diverse and complex video scenes.

\input{fig/fig_pipeline_dataset}
Our work advances video object removal in three key aspects:
(i) We introduce \textit{VOR}, a high-quality, large-scale hybrid dataset featuring diverse dynamic objects and complex multi-object scenarios across both camera-captured and synthesized environments.
(ii) We propose \textit{EffectErase}, a joint learning framework that integrates a Task-Aware Region Guidance module and an Effect Consistency loss to accurately identify and remove objects together with their visual effects.
(iii) We establish two benchmarks, \textit{VOR-Eval} and \textit{VOR-Wild}, providing a solid foundation for future research. The proposed method EffectErase achieves new state-of-the-art performance, surpassing existing methods in both quantitative metrics and visual quality.

%% file: fig/fig_limits_of_prev_inpainting.tex
\begin{figure}
\centering
\includegraphics[width=\linewidth]{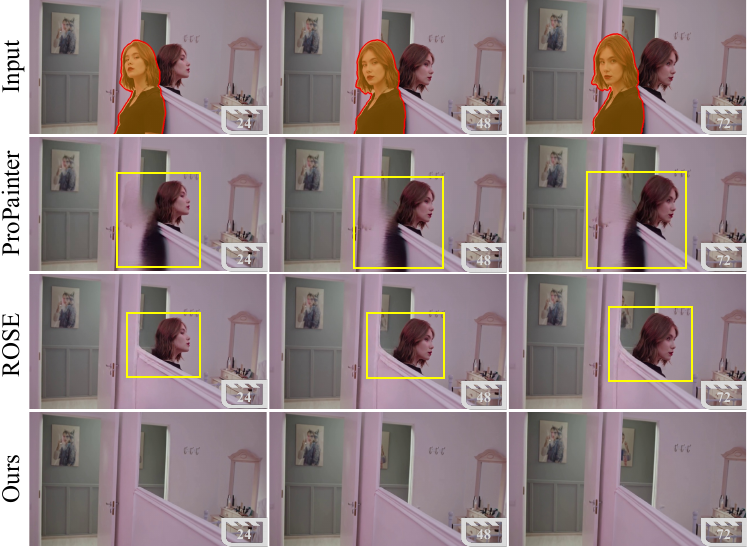}
\vspace{-7mm}
\caption{\textbf{Limitations of existing video object removal methods.}
While existing methods~\cite{zhou2023propainter,miao2025rose} can remove the main body within the input mask region, they often struggle to discover and remove the side effects (\eg, reflections) caused by the target object.
}
\label{fig_limits_of_prev_inpainting}
\vspace{-2mm}
\end{figure}

%% file: fig/fig_limits_of_prev_removal.tex
\begin{figure}
\centering
\includegraphics[width=\linewidth]{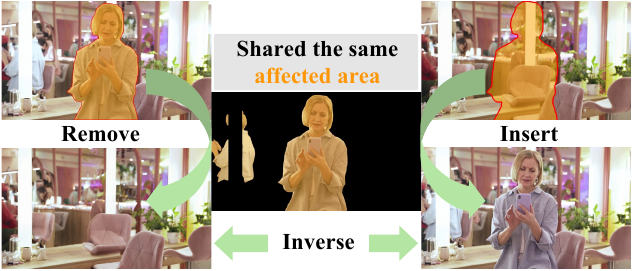}
\vspace{-0.66cm}
\caption{\textbf{Removal–Insertion.} Video object removal and insertion are inverse tasks that operate on the same affected regions.
}
\label{fig_remove_joint}
\vspace{-3mm}
\end{figure}

%% file: tab/dataset_comp.tex
\begin{table*}[t]
\centering
\small
\caption{Comparison of video object removal datasets. Our VOR dataset exceeds prior datasets in scale and diversity, offering broader object coverage and richer camera, object, and background dynamics. Further comparisons with image-level datasets are in supplementary.}
\vspace{-3mm}
\setlength\tabcolsep{5.3pt}
\resizebox{\textwidth}{!}{
\begin{tabular}{r|ccccccccccc}
\toprule
{Dataset} & {Source} & \makecell{{Dynamic} \\ {Camera}} &\makecell{{Dynamic} \\ {Object}} & \makecell{{Dynamic} \\ {Background}} & \makecell{{Scene} \\ {Types}} & \makecell{{Object} \\ {Classes}} & \makecell{{Image} \\ {Pairs}} & \makecell{{Video} \\ {Pairs}} & \makecell{{Average} \\ {Duration (s)}} & \makecell{{Total} \\ {Hours}} \\
\hline
RORD~\cite{Sagong_2022_BMVC}            & Real         & \xmarkg & \cmark & \xmarkg &  24 &\ \ 76  & \ \ \ \ 516.7K & \ 3,106 & \begray{–} &\ \ \ \ 5.98 \\
Video4Removal~\cite{wei2025omnieraser}   & Real         & \xmarkg & \cmark & \xmarkg &\ 6  & \ \ \begray{–}  &\  \ \ \ 134.3K & \begray{–} & \begray{–} & \ \ \ \ 1.55 \\
ROSE~\cite{miao2025rose}            & Synth.        & \cmark & \xmarkg & \xmarkg &  25 & 102  & \ 1{,}501.0K & 16{,}678 & 6.00 & \ \ 27.79 \\
\rowcolor{cyan!10}
\textbf{{VOR (Ours)}} & Real + Synth. & \cmark & \cmark & \cmark & \textbf{67} & \textbf{366}  &  \textbf{12,556.8K}& \textbf{60{,}000} & \textbf{8.72} & \textbf{145.33} \\
\bottomrule
\end{tabular}
}
\label{tab:remove_dataset}
\vspace{-3mm}
\end{table*}

%% file: fig/fig_pipeline_dataset.tex
\begin{figure*}[t]
    \centering
    \includegraphics[width=\textwidth]{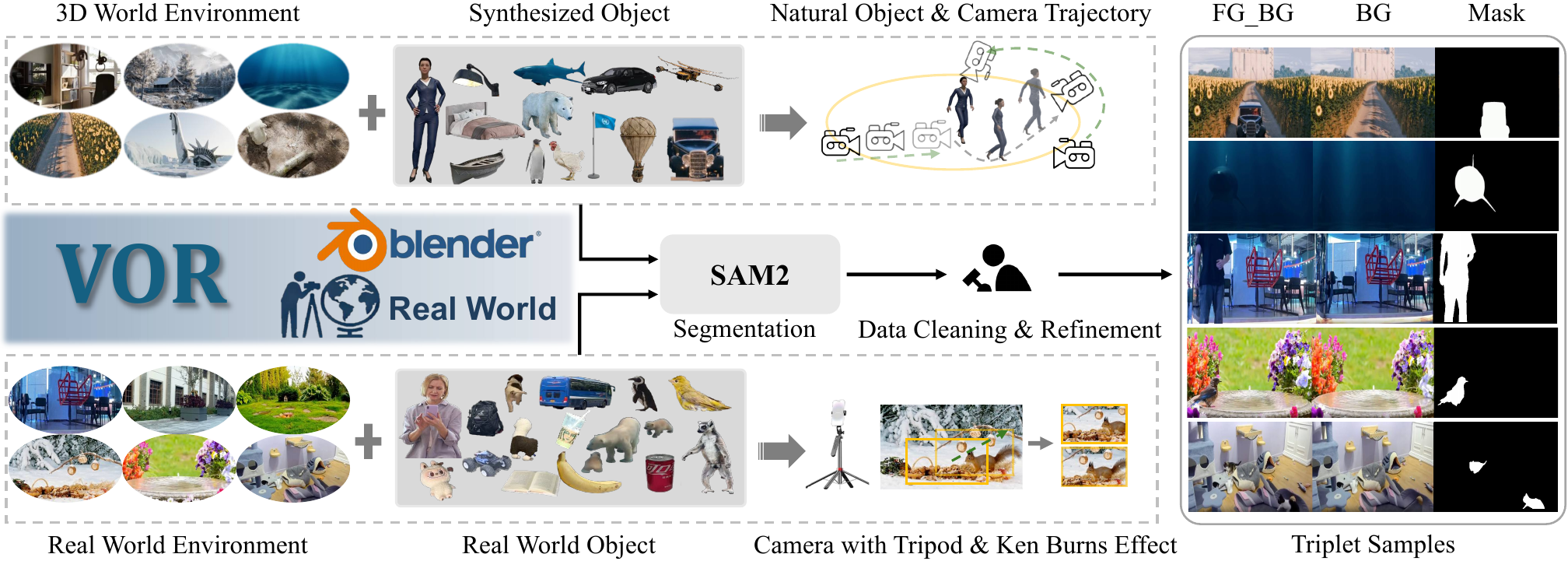
    }
    \vspace{-6mm}
    \caption{\textbf{Dataset Construction Pipeline of VOR.}
   VOR is a hybrid dataset combining synthetic data and real-world captures. Synthetic data are generated in Blender using 3D environments, objects, and animations collected from public sources, together with carefully designed natural object and camera trajectories. Real-world data are recorded across diverse scenes and object categories using cameras, followed by the Ken Burns effect to simulate camera motion. All videos are segmented by SAM2~\cite{ravi2024sam2} and manually cleaned and refined by human annotators. The final dataset comprises triplet pairs of videos with and without the target object, and the corresponding mask.
    }
    \label{fig:pipe_dataset}
    \vspace{-3mm}
\end{figure*}

%% file: sec/2_related.tex
\section{Related Work}
\label{sec:related}

%
\textbf{Video Inpainting} aims to reconstruct missing regions specified by a sequence of masks.
Early methods~\cite{wang2019video,chang2019free} use convolutional networks for spatiotemporal modeling but struggle with long-range propagation.
Subsequent works \cite{zhang2022flow,zhou2023propainter} exploit optical flow for additional motion cues. 
For example, ProPainter~\cite{zhou2023propainter} uses recurrent flow completion to improve controllability and temporal consistency.~To further enhance controllability, recent studies explore text-guided video inpainting by leveraging the priors of video diffusion models.~COCOCO~\cite{zi2025cococo}, for example, introduces motion capture to stabilize results.
Building on architectural advances, FloED~\cite{gu2024coherent} combines motion guidance with a multi-scale flow adapter to improve temporal consistency for removal and background restoration, while VideoPainter~\cite{bian2025videopainter} employs a lightweight context encoder to enhance background integration, foreground synthesis, and user control.
More recently, the unified video-synthesis baseline VACE~\cite{vace} introduces a context adapter with formalized temporal and spatial representations to support multiple tasks.
%
Despite these advances, existing inpainting models often overlook object effects, resulting in incomplete or visually inconsistent object removal.

\textbf{Object Removal} is a specialized form of inpainting that requires precise modeling of object-induced visual effects to achieve realistic results.
Early works primarily focus on image-level effects to ensure completeness and realism.
%
ObjectDrop~\cite{winter2024objectdrop} captures real scenes before and after removing a single object, but with limited scale.
SmartEraser~\cite{jiang2025smarteraser} and Erase Diffusion~\cite{liu2025erase} rely on synthetic datasets generated with segmentation~\cite{MOSE,MOSEv2} or matting, fail to reproduce realistic side effects such as shadows and reflections.
To improve realism, LayerDecomp~\cite{yang2025generative} and OmniPaint~\cite{yu2025omnipaint} construct costly camera-captured datasets. OmniPaint auto-labels unlabeled images with a model trained on limited real data, whereas RORem~\cite{li2025rorem} employs human annotators for refinement.
RORD~\cite{Sagong_2022_BMVC} and OmniEraser~\cite{wei2025omnieraser} mine static-camera videos to pair frames with and without the target, preserving natural effects, but remain limited to image-level removal and struggle in dynamic scenes.

\textbf{Video Object Removal} is more challenging, further requiring temporal consistency across frames beyond spatial fidelity.~Minmax-Remover~\cite{zi2025minimaxremovertamingbadnoise} simplifies a pre-trained video generator by discarding text inputs and cross-attention layers while distilling stage-1 outputs using a tailored minimax optimization objective.
However, this method only implicitly models video object effects and lacks access to a large and high-quality dataset.
ROSE~\cite{miao2025rose} introduces a synthesized dataset comprising multiple environments and approximately 27.8 hours of randomly captured video, along with a side-effect mask predictor.
However, its limited scale, omission of key effects such as deformation and dynamic object motion, and synthetic composition restrict generalization to real-world scenarios.

%% file: sec/3_method.tex
\section{Methodology}
\label{sec:meth}
%

%
\subsection{VOR Dataset}
\label{sec:meth:dataset}
\myparagraph{Overview.} 
As shown in~\cref{fig:pipe_dataset}, VOR is a hybrid dataset with two components: (1) {camera-captured videos} emphasizing physical realism and real-world distributions, and
(2) {synthesized videos} rendered with a 3D graphics engine to model dynamic cameras and multi-object interactions. 

\input{fig/fig_dataset_effect}

\myparagraph{Representative Object-Induced Effects.}
To better characterize object-induced effects under diverse conditions, as shown in~\cref{fig:effect_dataset}, we group them into five representative types: 
{{(1) Occlusion.}} This is the most common case where objects block parts of the scene. 
We further consider three subtypes based on transparency: opaque, semi-transparent (\eg, smoke), and transparent (\eg, glass), 
which pose different challenges for recovering occluded content from surrounding context.
{{(2) Shadow.}} Objects obstruct light, producing regions with varying intensity and shape. 
The main challenge lies in accurately localizing and inpainting these shadowed areas under diverse illumination.
{{(3) Lighting.}} Removing a light source changes scene brightness and color balance, requiring the model to estimate illumination effects on nearby regions and restore consistent lighting across frames.
(4) Reflection.~Objects are reflected on surfaces such as mirrors, water, or tiles. The model needs to disentangle and remove reflection artifacts while preserving the surface appearance.
{{(5) Deformation.}} Objects physically deform surrounding structures, \eg, curtains, grass, or nets. 
The model should recover the original geometry and texture with temporal coherence once the object is removed.

\myparagraph{Real-World Data.}  
%
We use fixed cameras to record paired videos that with and without target objects while keeping all other factors unchanged.~These videos are captured across diverse real-world  scenes, such as streets, parks, classrooms, rivers, and gyms, 
covering a wide range of static and dynamic objects, \eg, humans, animals, balls, and umbrellas.
The dataset spans different times of day and various weather conditions, \eg, sunny, cloudy, and rainy.

\input{fig/fig_pipeline_method}
\myparagraph{Synthesized Data.}  
{{(1) Diverse Scenes.}} We construct over 150 diverse 3D scenes from public repositories, covering a wide range of environments, weather, seasons, and full day lighting variations from morning to night.
{{(2) Objects and Motion.}} Unlike ROSE~\cite{miao2025rose}, where motion dynamics are solely induced by the camera, we curate common 3D objects and manually rig their motions, trajectories, and interactions.
We also design multi-object scenarios where only a subset of objects is removed, a setting largely overlooked in previous works.
{{(3) Multi-Camera Rendering.}} Rather than random trajectories, we design naturalistic multi-camera placements and motion paths to better approximate real-world cinematography and viewpoint diversity.

\myparagraph{Triplet Data Pairs.}  
{{(1) Camera Motion Simulation.}} 
For camera-captured pairs with and without the target object, we enrich motion diversity by applying the Ken Burns effect, combining smooth pans, zooms, and handheld head bob, following 14 predefined camera motion rules. We vary camera speed and trajectory within bounds so the moving window remains within the original frame. For each pair, five motion patterns are sampled from the 14 rules.
{{(2) Synthetic Data Combination.}} 
Given n objects and m camera configurations, we can construct (3$^\text{n}$~\!-~\!2$^\text{n}$)~\!$\times$~\!m pairs, substantially increasing both dataset scale and diversity.
{{(3) Mask Generation.}} To generate high-quality masks, we manually provide point prompts on key frames, verify the segmentation results, and propagate them across sequences using SAM2~\cite{ravi2024sam2} to obtain object masks sequences.
We then inspect each video segmentation result for data cleaning and manually refine the masks. 
Finally, by combining the validated masks with the video pairs, we construct triplet training data for subsequent learning.

\myparagraph{Data Statistics.} 
As summarized in~\Cref{tab:remove_dataset}, our dataset provides over 145 hours of video and \videonum paired videos, spanning 366 object classes and 443 different scenes. It substantially exceeds prior datasets in both scale and diversity, offering broader object coverage and richer variations in camera motion, object motion, and background dynamics.

\subsection{EffectErase}
\label{sec:meth:method}
\myparagraph{Overview.}
As shown in \cref{fig:pipe_methods}, the network encodes paired removal and insertion inputs with a pretrained VAE~\cite{kingma2013auto} and denoises the latents using a DiT~\cite{wan2025}.
On this backbone, our EffectErase incorporates three components:
1) Removal–Insertion Joint Learning, which trains both tasks together on the same affected regions and structural cues.
2) Task-Aware Region Guidance, which encodes object visual tokens and task-specific tokens to model spatiotemporal correlations between the object and its effects via cross attention, enabling flexible task switching;
3) Effect Consistency Loss, which enforces consistent effect regions between removal and insertion.

\myparagraph{Removal–Insertion Joint Learning.}
Most existing video object removal methods treat removal as an isolated task, often leading to insufficient awareness of affected regions and making it difficult to accurately localize and restore these areas. 
We propose a dual-learning paradigm in which removal and insertion share a common denoising backbone. 
Joint optimization of the two tasks provides complementary supervision, enabling the model to learn consistent affected regions and structural cues.
Specifically, video inputs are first encoded into the latent space using a pretrained VAE.
The video with objects $V^{o}$, the background video without objects $V^{b}$, and the corresponding mask ${M}$ are encoded into latent representations $\boldsymbol{x}^{o}$, $\boldsymbol{x}^b$, and $\boldsymbol{x}^{m}$, respectively.

To construct the noisy input $\boldsymbol{x}_{t}$ for diffusion training, a clean latent $\boldsymbol{x}$ obtained from the VAE is used, where $\boldsymbol{x} = \boldsymbol{x}^b$ for removal and $\boldsymbol{x} = \boldsymbol{x}^o$ for insertion. Random noise $\boldsymbol{z} \sim \mathcal{N}(0, I)$ is added through the forward process~\cite{esser2024scaling}:
\begin{equation}
\boldsymbol{x}_{t} = t\boldsymbol{x} + (1-t)\boldsymbol{z},
\end{equation}
where the timestep $t \in [0,1]$ is sampled from a logit-normal distribution. 
The denoising model $v_\theta$ is trained to predict the velocity $\boldsymbol{v} = \boldsymbol{x} - \boldsymbol{z}$ from the noisy latent $\boldsymbol{x}_{t}$, the timestep $t$, and the condition $\boldsymbol{c}$, with the objective defined as:
\begin{equation}
\mathcal{L}_\text{{denoise}} = \mathbb{E}_{\boldsymbol{z}, \boldsymbol{x}, t, \boldsymbol{c}}\big\| v_\theta(\boldsymbol{x}_{t}, t, \boldsymbol{c}) - \boldsymbol{v} \big\|^2,
\end{equation}
where the condition $\boldsymbol{c}$ guides the model to user-specified regions and differs across tasks: for {removal}, $\boldsymbol{c} = [\boldsymbol{x}^o ; \boldsymbol{x}^m]$; for {insertion}, $\boldsymbol{c} = [\boldsymbol{x}^b ; \boldsymbol{x}^f]$. Here $[\,;\,]$ denotes concatenation along the channel dimension and $\boldsymbol{x}^{f} = \boldsymbol{x}^{o} \odot \boldsymbol{x}^{m}$ with $\odot$ denoting element-wise multiplication.

To better fuse condition with noisy latents,  we introduce a lightweight adaptor $\mathcal{A}_{\phi}(\cdot)$ that combines $\boldsymbol{x}_{t}$ and $\boldsymbol{c}$:
\begin{equation}
\dot{\boldsymbol{x}}_{t} = \mathcal{A}_{\phi}([\boldsymbol{x}_{t};\boldsymbol{c}]).
\end{equation}

\myparagraph{Task-Aware Region Guidance.}
To model spatiotemporal correlations between the affected areas and objects and to support flexible switching between removal and insertion, we design a Task-Aware Region Guidance (TARG) module.
Task tokens $\boldsymbol{e}^\text{{task}}$ are extracted from a language model~\cite{raffel2020exploring}, while foreground tokens $\boldsymbol{e}^{{f}}$ are obtained by feeding a cropped foreground patch from a frame of $V^{{f}}=V^{{o}}\odot\mathcal{M}$ into the CLIP image encoder~\cite{radford2021learning}.
A lightweight projector $\mathcal{P}_{\psi}(\cdot)$ maps CLIP features into the token space.
The projected foreground embedding $ \mathcal{P}_{\psi}(\boldsymbol{e}^{{f}}) $ then replaces the placeholder token ``object'' in $\boldsymbol{e}^{\text{task}}$, forming a task-aware region representation:
\begin{equation}
\boldsymbol{e}^{\text{prompt}} = \boldsymbol{e}^\text{{task}}[\text{object}] \leftarrow \mathcal{P}_{\psi}(\boldsymbol{e}^{{f}}),
\end{equation}
which is injected into the backbone via cross-attention~\cite{vaswani2017attention} to guide the model in capturing spatiotemporal effect correlations between the object and its effects, enabling accurate localization of effect-related regions and flexible switching between removal and insertion.

\myparagraph{Effect Consistency Loss.}
Since video object removal and insertion are inverse operations, they share the same effect regions, covering both the object and its induced environmental changes.
{Under the joint-learning described above, the removal and insertion branches use different inputs and task tokens and therefore produce two sets of cross-attention maps. Because cross attention highlights effect-affected regions, we introduce an Effect Consistency (EC) loss to align the two branches, using insertion as auxiliary supervision for removal.
We collect cross-attention maps of each DiT block from both branches and max-pool across blocks to obtain $A^\text{{rm}}$ and $A^\text{{in}}$ for removal and insertion, respectively.}~A lightweight mapper $\mathcal{G}_{\omega}(\cdot)$ then projects them into soft affected region estimations:
\begin{equation}
{f}^\text{{rm}} = \mathcal{G}_{\omega}(A^\text{{rm}}), \quad
{f}^\text{{in}} = \mathcal{G}_{\omega}(A^\text{{in}}).
\end{equation}
As the implicitly learned affected areas may be unstable, we build a difference map prior ${f}^\text{diff}$ from the normalized distribution of the downsampled difference between $V^o$ and $V^b$. 
Unlike previous work~\cite{miao2025rose} that employs binary masks and loses change intensity information, such as variations in illumination and shadows, our soft distribution preserves detailed variations, better capturing the magnitude of the effects. 
{EC is computed once on the pooled maps, and gradients backpropagate through the mapper into all cross-attention layers, sharpening their focus on affected regions.}
The EC loss is formulated as:
\begin{equation}
\mathcal{L}_{\text{EC}} 
= \mathrm{KL}\!\left({f}^{\text{diff}} \,\|\, {f}^{\text{rm}}\right)
+ \mathrm{KL}\!\left({f}^{\text{diff}} \,\|\, {f}^{\text{in}}\right),
\end{equation}
which aligns effect regions across tasks and lets insertion provide complementary guidance for removal.

During training, the model is jointly optimized:
\begin{equation}
\mathcal{L}_\text{total} =
 \mathcal{L}_{\text{denoise}}^{\text{remove}}
+ \mathcal{L}_{\text{denoise}}^{\text{insert}}
+ \lambda \, \mathcal{L}_\text{{EC}},
\end{equation}
where the EC term is weighted by $\lambda$.
\input{tab/vor_rose_exp_comp}

%% file: fig/fig_dataset_effect.tex
\begin{figure}[t]
    \centering
    \includegraphics[width=0.47\textwidth]{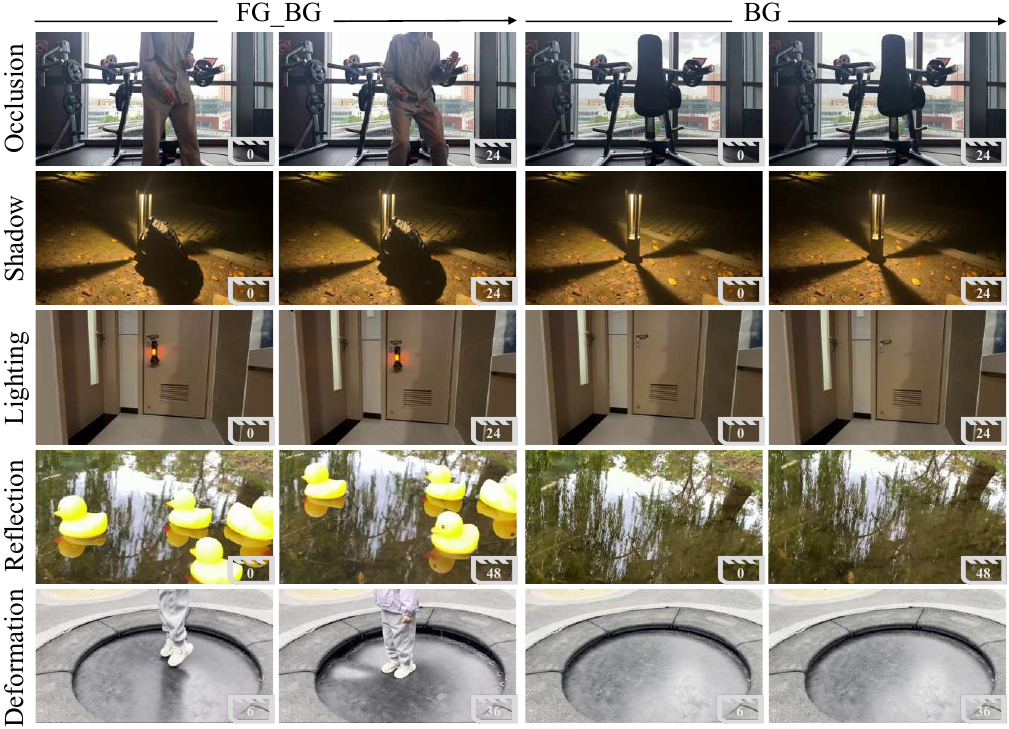}
    \vspace{-3.6mm}
    \caption{{Representative side effects in VOR dataset.}}
    \label{fig:effect_dataset}
    \vspace{-2mm}
\end{figure}

%% file: fig/fig_pipeline_method.tex
\begin{figure*}[t]
    \centering
    \includegraphics[width=\textwidth]{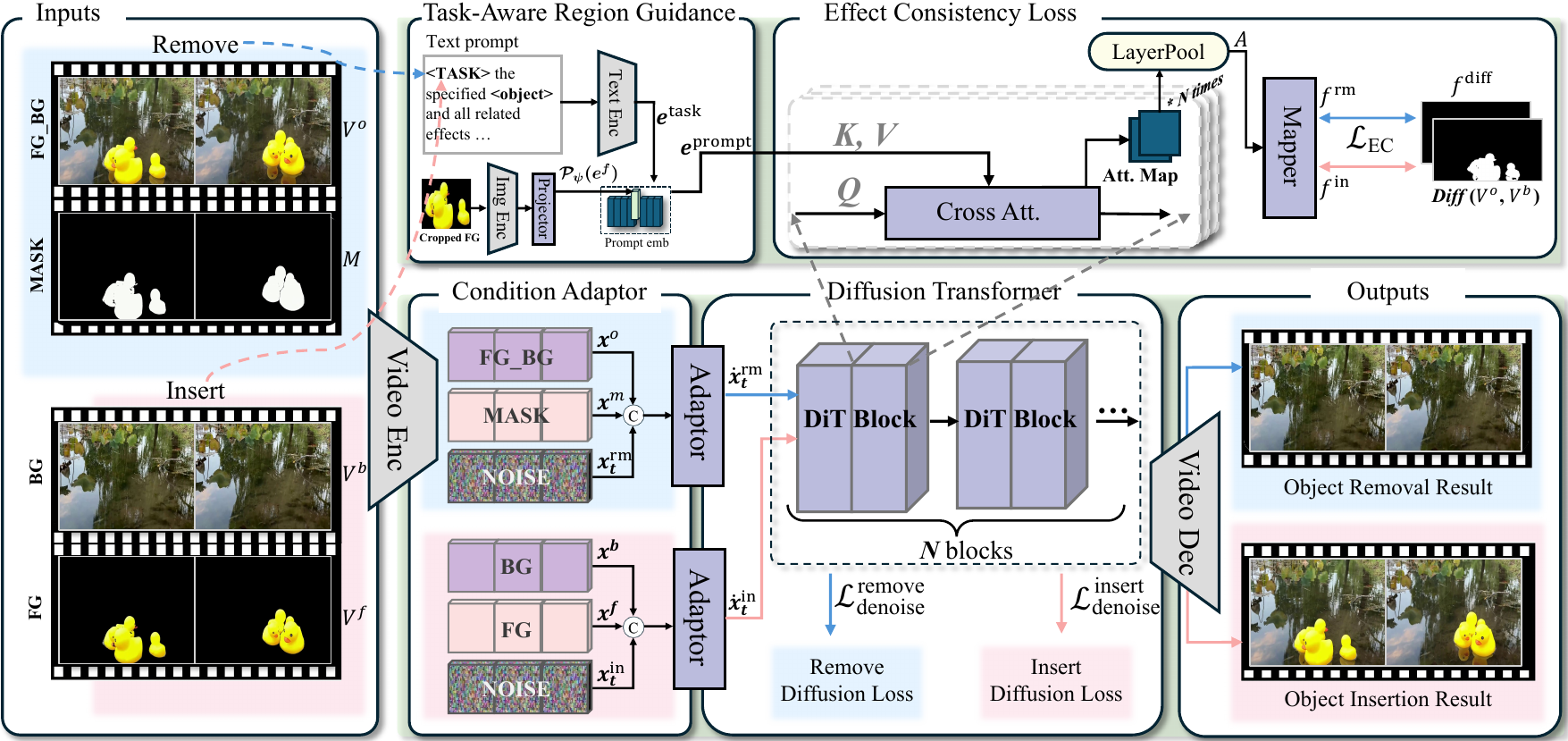}
    \vspace{-6mm}
    \caption{\textbf{The framework of EffectErase.} During training, removal and insertion pairs are encoded into the latent space by VAE and fused with noise via the Adaptor.
    Each DiT block performs cross-attention using the fused features $\dot{\boldsymbol{x}}_t$ as Query and $\boldsymbol{e}^{\text{prompt}}$ from Task-Aware Region Guidance as Key/Value, producing attention maps that highlight affected regions.
    We aggregate attention maps from all blocks and apply max pooling to obtain a maximal-activation map, which is supervised by the effect consistency loss $\mathcal{L}_{\text{EC}}$ to encourage both tasks to focus on the same affected area.
    At inference, users can flexibly switch the model between removal and insertion by modifying the inputs.
    }
    \label{fig:pipe_methods}
    \vspace{-3mm}
\end{figure*}

%% file: tab/vor_rose_exp_comp.tex
\begin{table*}[t]
\centering
\footnotesize
\caption{Quantitative results on ROSE and VOR. 
The best and second-best results are highlighted in bold and underlined, respectively.}
\label{tab:vor_results}
\vspace{-3mm}
\renewcommand{\arraystretch}{1.06}
\setlength{\tabcolsep}{6.5pt}
\begin{tabular}{l|cccc|cccc|cc}
\toprule
\multirow{2}{*}{Method} &
\multicolumn{4}{c|}{ROSE-Benchmark (with GT)} &
\multicolumn{4}{c|}{VOR-Eval (with GT)} &
\multicolumn{2}{c}{VOR-Wild (without GT)} \\
& PSNR↑ & SSIM↑ & LPIPS↓ & FVD↓ 
& PSNR↑ & SSIM↑ & LPIPS↓ & FVD↓ 
& QScore↑ &User↑ \\
\hline\hline
ObjectClear~\cite{zhao2025objectclear}                         & 29.535 & \underline{0.920} & \underline{0.076} & 742.829 & 22.583 & 0.787 & \underline{0.190} & 1391.858 & 8.979   & 4.75 \\ 
OmniPaint~\cite{yu2025omnipaint}                         & 27.569 & 0.910 & 0.085 & 809.645 & 21.511 & 0.781 & 0.201 & 1439.867  & 8.942  & 4.38\\ 
\hline
Propainter~\cite{zhou2023propainter}     & 27.200 & 0.915 & 0.095 &  171.020 & 21.975 & 0.800 & 0.225 & \ \ 589.012  & 8.860  & 4.88\\

DiffuEraser~\cite{li2025diffueraser}     & 26.502 & 0.898 & 0.128 & 167.483 & 21.946 & \underline{0.802} & 0.214 & \ \ 559.497  & 9.113  & 5.50\\

VACE~\cite{vace}                         & 20.805 & 0.694 & 0.174 & 254.117 & 17.677 & 0.591 & 0.294 & \ \ 806.476  & 8.229 & 1.50\\ 
\hline
MinMax-Remover~\cite{zi2025minimaxremovertamingbadnoise} & 26.770 & 0.905 & 0.099 & 137.840 & 21.963 & \underline{0.802} & 0.217 & \ \ 539.427  & 8.984  & 5.90\\
ROSE~\cite{miao2025rose}                 & \underline{31.122} & 0.917 & 0.077 & \ \ \underline{72.177} & \underline{22.966} & 0.792 & 0.203 & \ \ \underline{383.084}  & \underline{9.240}  & \underline{6.38}\\
\rowcolor{cyan!10}\textbf{EffectErase (Ours)}             & \textbf{32.161} & \textbf{0.948} & \textbf{0.039} & \ \ \textbf{55.578}
                                         & \textbf{23.750} & \textbf{0.806} & \textbf{0.170} & \textbf{\ \ 342.871}   & \textbf{9.280} & \textbf{7.20}\\
\bottomrule
\end{tabular}
\end{table*}

%% file: sec/4_exp.tex
\section{Experiments}
\label{sec:exp}

\myparagraph{Implementation.} Our method is built on the Wan 2.1~\cite{wan2025wan} video generation model and fine-tuned with LoRA~\cite{hu2022lora} on the VOR dataset.
The input resolution is set to $832 \times 480$, and 81 consecutive frames are randomly sampled for training.
The model is trained for 120K iterations with a total batch size of 8 on 8 H100 GPUs, using a learning rate of $1 \times 10^{-5}$ and a LoRA rank of 256. All results are generated with 50 denoising steps.

\input{fig/fig_result_comp}
\input{fig/fig_result_comp_wild}
\input{fig/fig_result_insert}
\myparagraph{Evaluation Data.} We evaluate EffectErase against existing methods on three datasets:  
(1) \textit{ROSE-Benchmark}, a synthetic dataset that provides paired videos for object removal evaluation;
(2) \textit{VOR-Eval}: the test split of our VOR dataset described in~\cref{sec:meth:dataset}, which contains 43 paired videos. 
(3) \textit{VOR-Wild}: a test set consisting of 195 diverse real-world videos collected from the internet, featuring dynamic objects and their associated effects.
%

%
\myparagraph{Evaluation Metrics.} For datasets with ground truth (ROSE and VOR-Eval), we adopt standard fidelity metrics, including PSNR~\cite{hore2010image}, SSIM~\cite{wang2004image}, LPIPS~\cite{zhang2018unreasonable}, and FVD~\cite{unterthiner2019fvd}.
For VOR-Wild, which lacks ground truth, we conduct a user study where 20 volunteers rate the results, and further introduce Qscore, a metric that leverages the Qwen-VL model~\cite{bai2025qwen2} to assess the quality of generated videos based on removal completeness and visual artifacts.

\subsection{Comparison with State-of-the-Art Methods.} We compare EffectErase with several state-of-the-art image inpainting methods~\cite{zhao2025objectclear,yu2025omnipaint} applied in a per-frame manner, video inpainting methods~\cite{vace,zhou2023propainter,li2025diffueraser}, and advanced video object removal methods~\cite{zi2025minimaxremovertamingbadnoise,miao2025rose}.

\myparagraph{Quantitative Evaluation.} As shown in~\Cref{tab:vor_results}, current image inpainting methods~\cite{zhao2025objectclear,yu2025omnipaint} operate on individual frames using 2D models without temporal modeling, and therefore fail to maintain temporal consistency in videos.
Recent video inpainting methods~\cite{zhou2023propainter,vace,li2025diffueraser} do not explicitly model object side effects, resulting in unnatural removal outcomes.
Existing video object removal methods~\cite{miao2025rose,zi2025minimaxremovertamingbadnoise} lack spatiotemporal correlation modeling between the object and its side effects, and consequently often produce artifacts and residual traces of the removed objects.
Overall, EffectErase achieves state-of-the-art performance across all datasets and evaluation metrics. 
It obtains the best scores on the video quality metric FVD, demonstrating superior temporal smoothness and consistency of the generated videos.
Our method also achieves the highest QScore and user feedback ratings, further demonstrating its effectiveness in producing visually convincing removal results.

\myparagraph{Qualitative Evaluation.} Qualitative comparisons are  presented  in~\cref{fig:res_comp} and~\cref{fig:res_comp_wild}.  
Video inpainting methods~\cite{vace,zhou2023propainter} often produce artifacts in masked regions and fail to completely remove the side effects caused by the removed objects.
Previous object removal approaches, such as ROSE~\cite{miao2025rose} and MinMax-Remover~\cite{zi2025minimaxremovertamingbadnoise}, perform well in removing target objects but still struggle with side effects, especially in occlusion, shadow, lighting, reflection and deformation scenarios.  
In contrast, EffectErase effectively removes both target objects and their associated effects, resulting in clean, coherent, and high-quality outcomes.  

\input{tab/ablation}
\subsection{Ablation Studies}

\myparagraph{Effectiveness of Consistency Loss.} The proposed EC loss encourages removal and insertion to focus on the same side-effect regions, strengthening the model’s attention to affected areas.
As shown in the~\Cref{tab:ablation_vor}, adding the EC loss consistently improves the baseline across all metrics, with FVD decreasing from 368.664 to 354.545.

\myparagraph{Effectiveness of Task-Aware Region Guidance.} The TARG module captures spatiotemporal correlations between objects and their side effects, enabling the model to localize and perceive affected regions.
As shown in~\Cref{tab:ablation_vor}, TARG enables the model to produce higher-quality erasure results, with SSIM improving significantly from 0.737 to 0.780, validating the effectiveness of this design.

\myparagraph{Effectiveness of Synthesized Data.} Incorporating high-quality synthesized data increases data diversity and exposes the model to a broader range of appearance variations and motion patterns.
As shown in~\Cref{tab:ablation_vor},  training with both real and synthetic data leads to noticeably better generalization on VOR-Eval, producing cleaner backgrounds and more stable temporal restoration. 
This mixed training setup yields consistent improvements across metrics, with LPIPS decreasing markedly from 0.193 to 0.170.

\subsection{More Applications.} EffectErase can be directly adapted to object insertion by simply modifying the task prompt without additional training.  
As shown in~\cref{fig:res_comp_insert}, the model synthesizes realistic object side effects even when only the target objects are specified. 
In the first two rows, EffectErase generates realistic shadows for inserted dynamic objects such as a leaf and a traffic cone, while the third row shows its ability to produce natural light reflections on glossy ceramic tiles.

%% file: fig/fig_result_comp.tex
\begin{figure*}[t]
    \centering
    \includegraphics[width=\linewidth]{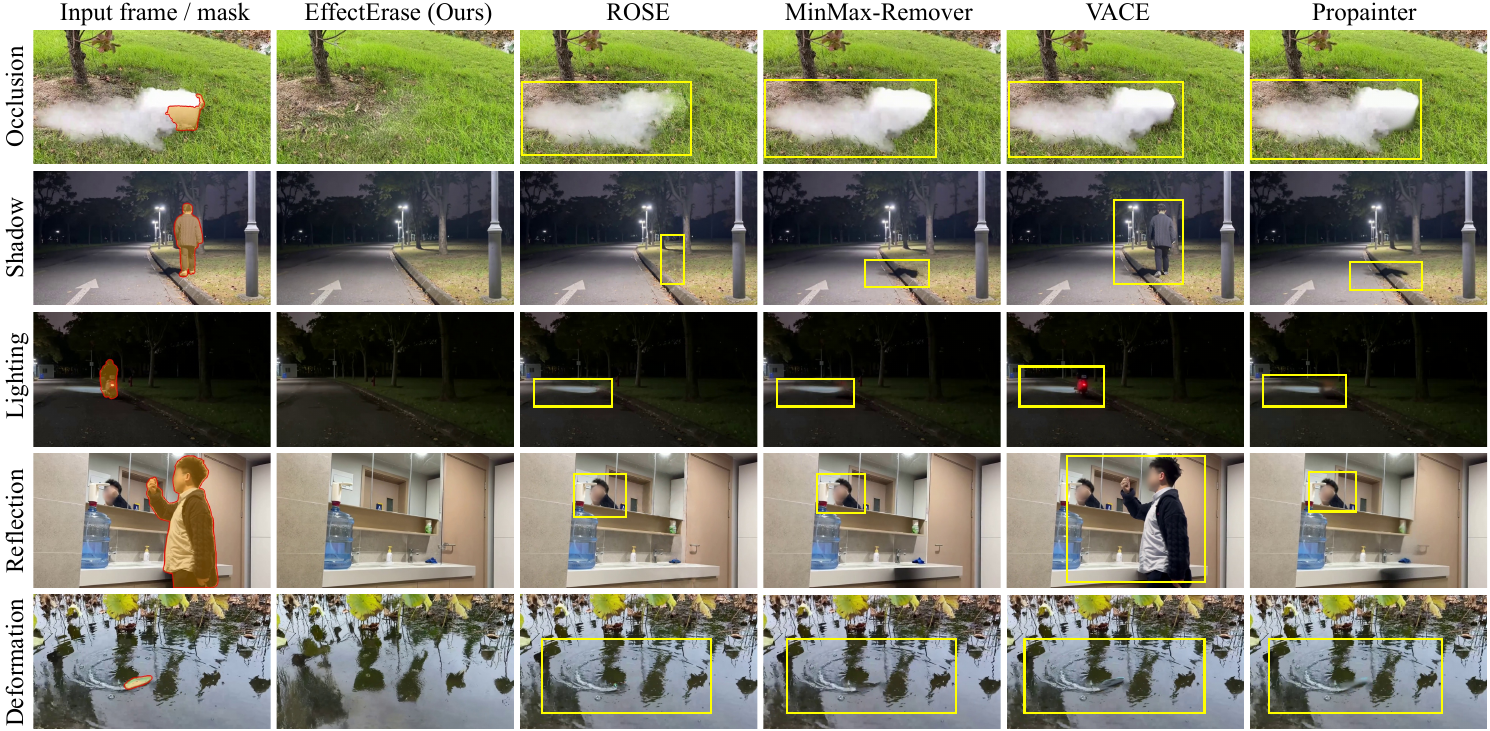}
    \\
    \vspace{-3mm}
    \caption{{Qualitative results on VOR-Eval.}
    Inpainting models (VACE~\cite{vace}, Propainter~\cite{zhou2023propainter}) fail to erase effects beyond the mask, while removal models (ROSE~\cite{miao2025rose}, MinMax-Remover~\cite{zi2025minimaxremovertamingbadnoise}) leave artifacts. EffectErase effectively removes the target objects and their effects.
    }
    \label{fig:res_comp}
\end{figure*}

%% file: fig/fig_result_comp_wild.tex
\begin{figure*}[t]
    \centering
    \includegraphics[width=\linewidth]{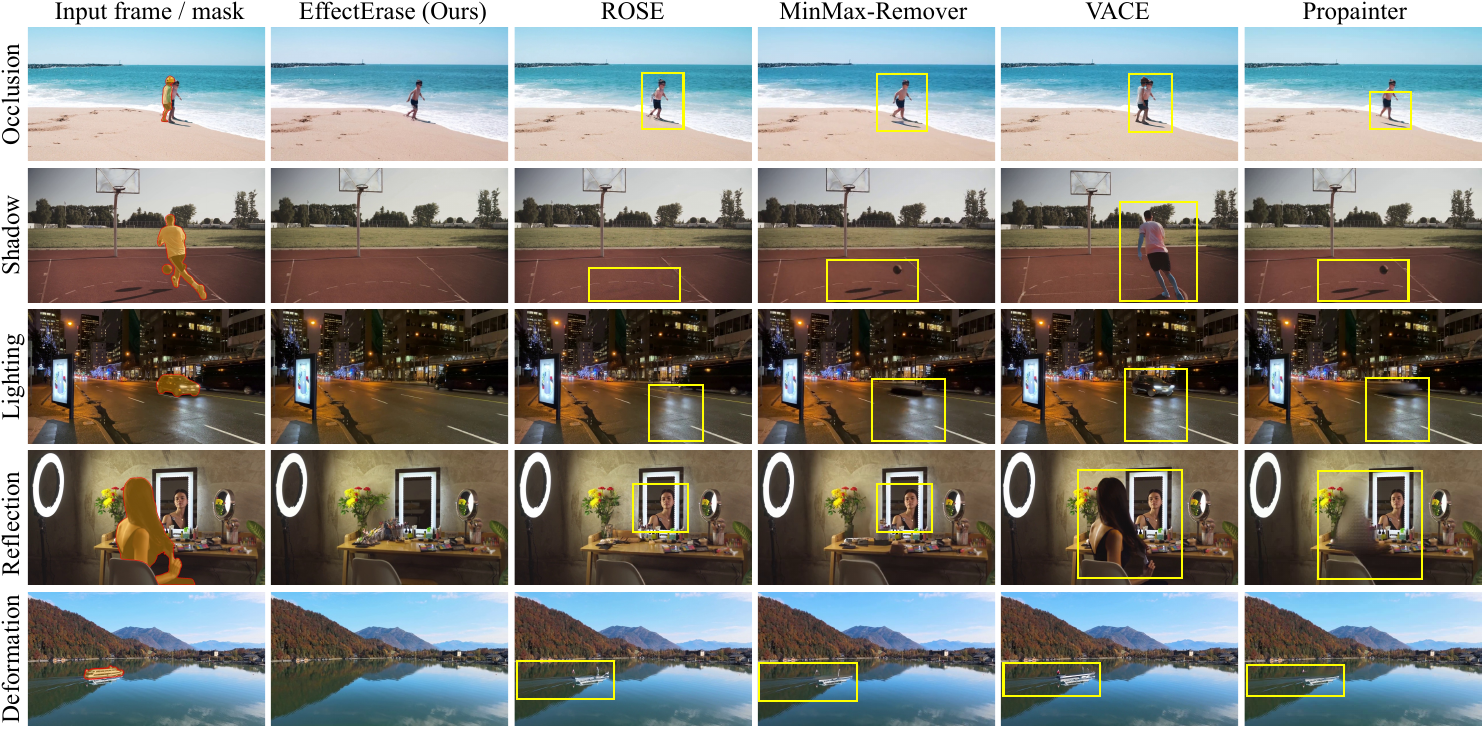}
    \vspace{-6.76mm}
    \caption{{Qualitative results on VOR-Wild.}
    %
  EffectErase remains robust across in-the-wild scenarios such as multi-person occlusions, fast-moving sports, nighttime headlights, mirror reflections, and open-water boat scenes. Best viewed zoomed in.
    }
    \label{fig:res_comp_wild}
    \vspace{-2mm}
\end{figure*}

%% file: fig/fig_result_insert.tex
\begin{figure}[t]
    \centering
    \includegraphics[width=0.992\columnwidth]{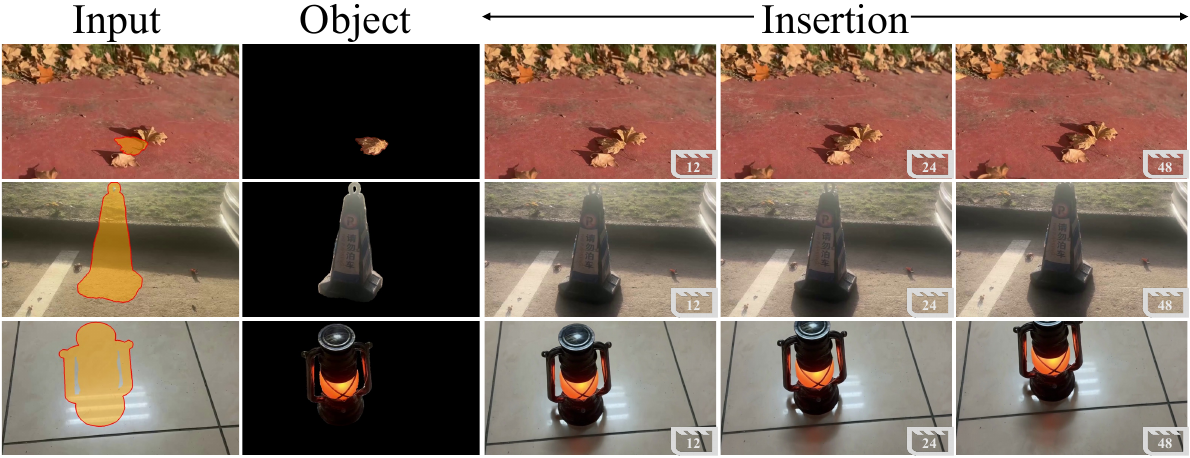}
    \vspace{-6.6mm}
    \caption{\textbf{Video Object Insertion by EffectErase.}
EffectErase seamlessly adapts to insertion, preserving background content while naturally integrating dynamic objects with realistic object-induced effects, \eg, shadows and reflections.
    }
    \vspace{-3mm}
    \label{fig:res_comp_insert}
\end{figure}

%% file: tab/ablation.tex
\begin{table}[t]
\centering
\caption{Ablation study on VOR-Eval. Based on VOR real-world data (Real), the removal performance improves progressively by adding the consistency loss ($\mathcal{L}_\text{{EC}}$), Task-Aware Region Guidance (TARG), and synthesized training data (Syn.).}
\label{tab:ablation_vor}
\vspace{-3mm}
\footnotesize
\renewcommand{\arraystretch}{1.1}
\setlength{\tabcolsep}{3pt}
\begin{tabular}{lcccc|cccc}
\hline
Exp. & Real & $\mathcal{L}_\text{{EC}}$ & TARG & Syn.  & PSNR$\uparrow$ & SSIM$\uparrow$ & LPIPS$\downarrow$ & FVD$\downarrow$ \\
\hline\hline
(a) & \cmark&  &  &  & 20.409  & 0.720 & 0.243& 368.664 \\
(b) & \cmark& \cmark&  &  & 21.020 & 0.737 & 0.224 & 354.545 \\
(c) & \cmark& \cmark& \cmark&  & 23.101 & 0.780 & 0.193 & 349.094 \\
\rowcolor{cyan!10}(d) & \cmark& \cmark& \cmark& \cmark& \textbf{23.750} & \textbf{0.806} & \textbf{0.170} & \textbf{342.871} \\
\hline
\end{tabular}
\vspace{-0.5cm}
\end{table}

%% file: sec/5_con.tex
\section{Conclusion}
\label{sec:con}

We address the challenging effect-aware video object removal by introducing the VOR dataset and EffectErase framework. VOR is a large hybrid dataset consisting of camera-captured and synthesized videos, covering common categories of object-induced effects, with two evaluation benchmarks VOR-Eval and VOR-Wild. Building on VOR, we propose EffectErase to jointly learn video object removal and insertion. EffectErase leverages Task-Aware Region Guidance to model spatiotemporal object–effect correlations, and enforces an Effect Consistency loss to align effect regions across tasks. Extensive experiments and ablations validate the contribution of each component. EffectErase achieves state-of-the-art performance, delivering high-quality removal of objects and their effects in complex scenes, and naturally extends to realistic object insertion.

\noindent\textbf{Limitation.}~EffectErase requires an input mask to specify the removal region, and a future direction is to support more user-friendly interactions, \eg, text and speech.

%% file: sec/X_suppl.tex
\setcounter{section}{0}
\setcounter{figure}{0}
\setcounter{table}{0}
\renewcommand{\thesection}{\Alph{section}}
\renewcommand{\thetable}{\Roman{table}}
\renewcommand{\thefigure}{\Roman{figure}}
\renewcommand{\theequation}{\roman{equation}}

\twocolumn[
    \begin{center}
        \vspace{1.6em}
        \textbf{\Large{Supplementary Material for EffectErase}}\\
        \vspace{3.6em}
    \end{center}
]

In the supplement, we provide additional dataset details in \cref{sec:more_dataset}, further method descriptions in \cref{sec:more_method}, and more qualitative results in \cref{sec:more_results}.

\section{Details of Dataset Construction}
\label{sec:more_dataset}
In this section, we provide a detailed description of the captured and rendered components of our \textbf{V}ideo \textbf{O}bject \textbf{R}emoval (\textbf{VOR}) dataset used to train EffectErase.
\subsection{Real-World Data}
\myparagraph{Consistent Data Pairs.}
Each pair consists of one video where the target object is present with its effects and a counterpart where both are absent. 
To keep the two recordings identical, as shown in \cref{fig:ios}, we develop a custom capture app that locks exposure and focus across the entire pair, ensures matched file names and fixed recording durations, enables Bluetooth triggering to avoid screen-touch motion, and uses a tripod to eliminate camera shake.

\myparagraph{Diverse Scenes and Objects.}
We collect data across a wide range of real-world environments, including parks, campuses, and streets, spanning a total of 293 scenes and covering over 45 scene categories. 
The dataset also features a broad set of objects, ranging from static items such as sports balls and tools to dynamic subjects including children, teenagers, and various vehicles.

\myparagraph{Ken Burns Effects.}
We propose an extended version of the Ken Burns effect that provides fourteen distinct camera-motion patterns. 
These include basic zoom-in and zoom-out motions; directional motions such as panning left or right and tilting up or down; combined zoom–translation motions; a walk-bob motion that mimics the vertical sway of handheld footage; and a random-combo mode that randomly mixes zoom and translation directions.
For each clip, we randomly select five motion types and assign each type a randomized zoom curve and translation intensity. 
The module then updates a virtual camera center over time and crops the corresponding view to a fixed resolution, producing natural and diverse camera-movement variants that enhance training for the video object removal model.
\subsection{Synthesized Data}
\myparagraph{3D Enviroments.}
We collect 150 high-quality 3D environment assets from free online resources. These scenes cover a wide range of realistic daily-life settings across both indoor and outdoor domains, \eg city streets, farms, coastal areas, mountains, parking lots, classrooms and forests.

\myparagraph{Characters with Animations.}
We include a diverse set of animated characters and objects, such as dancing humans, walking bears, moving boats, and flying balloons, covering realistic, anime, and game-style visual domains.

\myparagraph{Camera Trajectories.}
Due to the wide variety of camera motions and shooting angles in real scenarios, we aim to cover as many camera movement patterns as possible. To this end, we manually design both realistic camera paths and natural camera motion behaviors such as zoom and pan, thereby ensuring that the synthesized movements closely mimic human-operated filming practices.
\input{fig/fig_ios}
\input{tab/all_dataset_comp}
\subsection{Mask Annotation}
We first provide a point prompt to obtain the mask in the first frame and manually verify its quality. 
The same point prompt is then fed to SAM2~\cite{ravi2024sam2} to propagate the mask across the entire sequence. 
We review all propagated mask sequences and remove those that fail to maintain stable and complete object coverage across all frames.

\subsection{Dataset Statics}
As shown in ~\Cref{tab:all_remove_dataset}, we provide a detailed comparison between our VOR dataset and existing image- and video-based removal datasets. 
We summarize the image-based datasets and the video-based datasets.
Compared with prior work, VOR offers substantially richer scene diversity, broader object coverage, longer video durations, and a significantly larger number of paired sequences.

Since no unified scene taxonomy exists across datasets, we introduce our own categorization scheme to standardize all scene types in~\cref{fig:tree}, covering both indoor and outdoor environments with a total of 67 comprehensive categories. Specifically, for RORD~\cite{Sagong_2022_BMVC}, its original scene labels are merged into our taxonomy; for Video4Removal~\cite{wei2025omnieraser}, scene types are assigned based on the descriptions in the paper and aligned with our scheme; and for ROSE~\cite{miao2025rose}, we manually inspect every scene in the raw data and annotate them according to our proposed categorization.

For video pair counts, the numbers for RORD~\cite{Sagong_2022_BMVC} are obtained by counting the lowest-level folders in the dataset structure. The total video hours of RORD~\cite{Sagong_2022_BMVC} and Video4Removal~\cite{wei2025omnieraser} are estimated by converting the total frame count to duration using 24~fps.

\input{fig/fig_tree}

\section{Method Details}
\label{sec:more_method}
\subsection{Details of the Proposed modules}
\myparagraph{Adaptor Details.}
The adaptor is implemented as a 3D convolutional layer with a kernel size of \(1\times2\times2\) and a stride of \(1\times2\times2\). 
To improve convergence, the first sixteen input channels of its weights are copied from the convolution used in the original patch-embedding module, while the remaining channels are initialized with Xavier uniform initialization~\cite{glorot2010understanding}. All bias terms are zero-initialized.
\input{fig/fig_prompt}
\myparagraph{Projector Details.}
The projector maps the object-image features extracted by the image encoder into the latent space required by our model. It is composed of two sequential MLP blocks: the first transforms the input embedding dimension to the output dimension, and the second further refines the representation with a residual MLP. 
Each block applies LayerNorm~\cite{ba2016layer}, a linear projection, a GELU activation~\cite{hendrycks2016gaussian}, and a second linear projection, while the second block includes a residual connection. 
A final LayerNorm is applied to stabilize the projected token.

\myparagraph{Mapper Details.}
The mapper predicts an effect-area distribution map from the fused cross-attention features. 
We aggregate the cross-attention maps from all DiT layers~\cite{Peebles2022DiT} and apply max-pooling across layers to obtain a compact feature volume. This volume is then processed by the mapper, implemented as a lightweight per-pixel MLP operating on the channel dimension.
The module applies a linear projection, a GELU activation~\cite{hendrycks2016gaussian}, and a second linear projection to produce a logit map for each frame.

\subsection{Training details}
Similar to previous work~\cite{miao2025rose}, the backbone model is a controllable generation variant of Wan2.1 1.3B~\cite{wan2025wan}. We optimize the network with AdamW~\cite{adamw} using a learning rate of \(1\times10^{-4}\) and a batch size of 1 through gradient accumulation. Training is conducted for up to 120K iterations. To adapt the base model to the video object-removal task, we apply LoRA~\cite{hu2022lora} to the attention projections \(q, k, v, o\) and the feed-forward layers \texttt{ffn.0} and \texttt{ffn.2}, with all LoRA weights initialized using Kaiming initialization~\cite{he2015delving}.
\subsection{Inference details}
During inference, the model supports both removal and insertion. 
For removal, we provide the input video together with a mask video, and the model outputs the object-removed result. 
For insertion, we provide the background video and an object video, and the model generates the inserted output. 
All denoising steps are set to 50.

\subsection{Metric details}
\myparagraph{QScore.}
To further assess the removal quality, we use the Qwen-VL model~\cite{bai2025qwen2} to evaluate each removed video with a designed prompt as shown in ~\cref{fig:prompt}. The evaluation considers both removal completeness and visual artifacts, and the final QScore is obtained by averaging the results.

\myparagraph{User Study.}
We conduct a user study with 20 volunteers, where each participant scores 195 generated videos from VOR-Wild, and the final score is obtained by averaging all individual ratings across participants.

\section{More Results}
\label{sec:more_results}
\subsection{Effect-region Erasing Evaluation}
As shown in the \cref{tab:region}, EffectErase effectively removes effect regions outside the object mask, with evaluation metrics computed only over the corresponding effect regions.
 \vspace{-1.5em}
\begin{table}[h]
\centering
\caption{Effect-region erasing evaluation.}
 \vspace{-0.8em}
\label{tab:region}
\setlength\tabcolsep{12.6pt}
    \resizebox{0.98\linewidth}{!}{
\scriptsize
\begin{tabular}{l|cccc}
\toprule
Method & PSNR ↑ & SSIM ↑ & LPIPS ↓ & FVD ↓ \\
\hline
ROSE        &30.267 &  0.930 & 0.084 & 135.013   \\
\rowcolor[HTML]{E1F4FC}
\textbf{EffectErase} &\textbf{32.747} & \textbf{0.939} &  \textbf{0.069}&  \ \ \textbf{98.266}  \\
\bottomrule
\end{tabular}

   }
    \vspace{-1.2em}
\end{table}

\subsection{More Results of the Insertion Task}
Please refer to ~\cref{fig:insert_more} for additional results of EffectErase applied to the insertion task.
\input{fig/fig_insert_more}

\subsection{More Results of EffectErase}
Please refer to ~\cref{fig:comp_more} for additional results of EffectErase on in-the-wild data.
\input{fig/fig_com_more}

\subsection{More Comparison with SOTA Methods}
Please refer to ~\cref{fig:comp_vor} for additional qualitative comparisons with state-of-the-art methods. 
\input{fig/fig_comp_vor}

\subsection{Failure Cases and Analysis}
\begin{figure}[H]
  \centering
  \vspace{-3mm}
  \includegraphics[width=\linewidth]{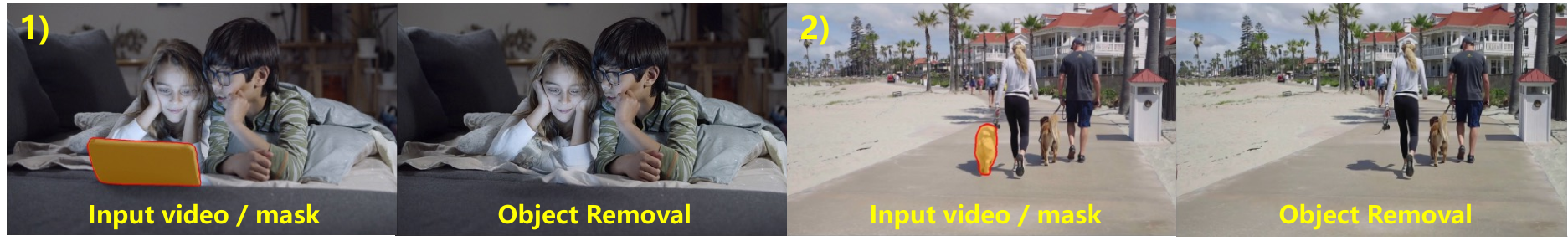}
  \vspace{-7mm}
  \caption{Failure cases when effects or accessories cannot be clearly attributed to the target object.}
  \label{fig:failurecase}
\end{figure}

Failure cases mainly arise when it is ambiguous whether effects or accessories belong to the target object. As shown in the ~\cref{fig:failurecase}, 1) the residual lighting may originate from other light sources, yet remains visually natural after removal; 2) parts of the dog’s shadow are heavily entangled with the person’s shadow, and the leash cannot be clearly assigned to either the dog or the person.

\vspace{20mm}

%% file: fig/fig_ios.tex
\begin{figure}[t]
    \centering
    \includegraphics[width=0.992\columnwidth]{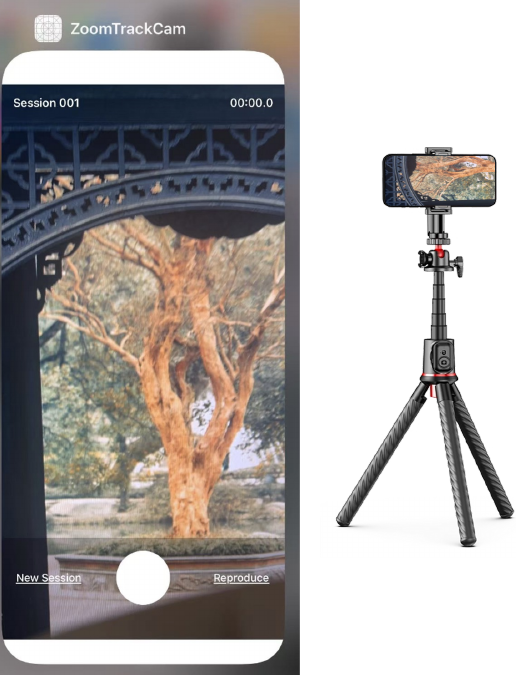}
\caption{\textbf{Data capture software.} Our app records aligned video pairs by locking exposure and focus, matching file names and durations, enabling reliable Bluetooth triggering for stable control, and using a tripod to remove camera shake.}
    \label{fig:ios}
\end{figure}

%% file: tab/all_dataset_comp.tex
\begin{table*}[t]
\centering
\small
\caption{Comparison of object removal datasets. Image-level datasets are listed above the line, and video-level datasets are listed below. “–” denotes unreported or not applicable. Synth.~(3D) denotes data generated using a graphics rendering engine, while Synth.~(paste) denotes data created by directly pasting cropped foreground objects onto backgrounds.}
\vspace{-3mm}
\setlength\tabcolsep{5.3pt}
\resizebox{\textwidth}{!}{
\begin{tabular}{r|ccccccccccc}
\toprule
{Dataset} & {Source} & 
\makecell{{Dynamic} \\ {Camera}} &
\makecell{{Dynamic} \\ {Object}} &
\makecell{{Dynamic} \\ {Background}} &
\makecell{{Scene} \\ {Types}} &
\makecell{{Object} \\ {Classes}} &
\makecell{{Image} \\ {Pairs}} &
\makecell{{Video} \\ {Pairs}} &
\makecell{{Average} \\ {Duration (s)}} &
\makecell{{Total} \\ {Hours}} \\
\hline
ObjectDrop~\cite{winter2024objectdrop}  & Real & \xmarkg & \xmarkg & \xmarkg & \begray{–} & \begray{–} & 2.5K & \begray{–} & \begray{–} & \begray{–} \\
Syn4Removal~\cite{jiang2025smarteraser} & Synth. (paste) & \xmarkg & \xmarkg & \xmarkg & \begray{–} & \begray{–} & 1{,}000K & \begray{–} & \begray{–} & \begray{–} \\
LayerDecomp~\cite{yang2025generative}   & Synth. (paste) & \xmarkg & \xmarkg & \xmarkg & \begray{–} & \begray{–} & 6.0K & \begray{–} & \begray{–} & \begray{–} \\
OmniPaint~\cite{yu2025omnipaint}        & Real & \xmarkg & \xmarkg & \xmarkg & \begray{–} & \begray{–} & 3.3K & \begray{–} & \begray{–} & \begray{–} \\
RORem~\cite{li2025rorem}                & Synth. (paste) & \xmarkg & \xmarkg & \xmarkg & \begray{–} & \begray{–} & 201.1K & \begray{–} & \begray{–} & \begray{–} \\
\Xhline{2\arrayrulewidth} 
RORD~\cite{Sagong_2022_BMVC}            & Real  & \xmarkg & \cmark & \xmarkg &  24 &  76 & 516.7K & 3{,}106 & \begray{–} & 5.98 \\
Video4Removal~\cite{wei2025omnieraser}  & Real  & \xmarkg & \cmark & \xmarkg &   6 & \begray{–} & 134.3K & \begray{–} & \begray{–} & 1.55 \\
ROSE~\cite{miao2025rose}                & Synth.(3D) & \cmark & \xmarkg & \xmarkg & 25 & 102 & 1{,}501.0K & 16{,}678 & 6.00 & 27.79 \\
\rowcolor{cyan!10}
\textbf{VOR (Ours)} & Real + Synth.(3D) & \cmark & \cmark & \cmark & \textbf{67} & \textbf{366} & \textbf{12{,}556.8K} & \textbf{60{,}000} & \textbf{8.72} & \textbf{145.33} \\
\bottomrule
\end{tabular}
}
\label{tab:all_remove_dataset}
 
\end{table*}

%% file: fig/fig_tree.tex
\begin{figure}[t]
    \centering
    \includegraphics[width=0.992\columnwidth]{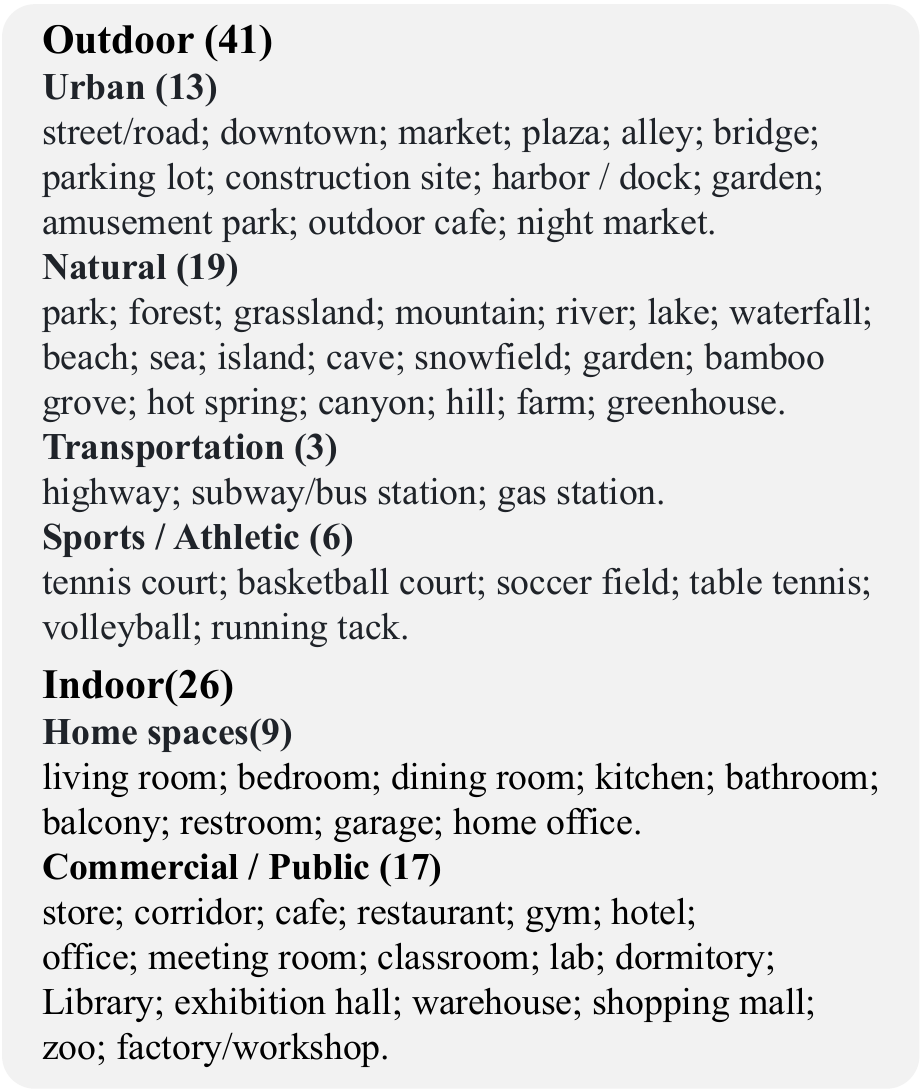}
\caption{\textbf{Scene category hierarchy.} Our taxonomy organizes 67 scene types into structured outdoor and indoor groups.}
    \label{fig:tree}
\end{figure}

%% file: fig/fig_prompt.tex
\begin{figure*}[]
    \centering
    \includegraphics[width=0.992\textwidth]{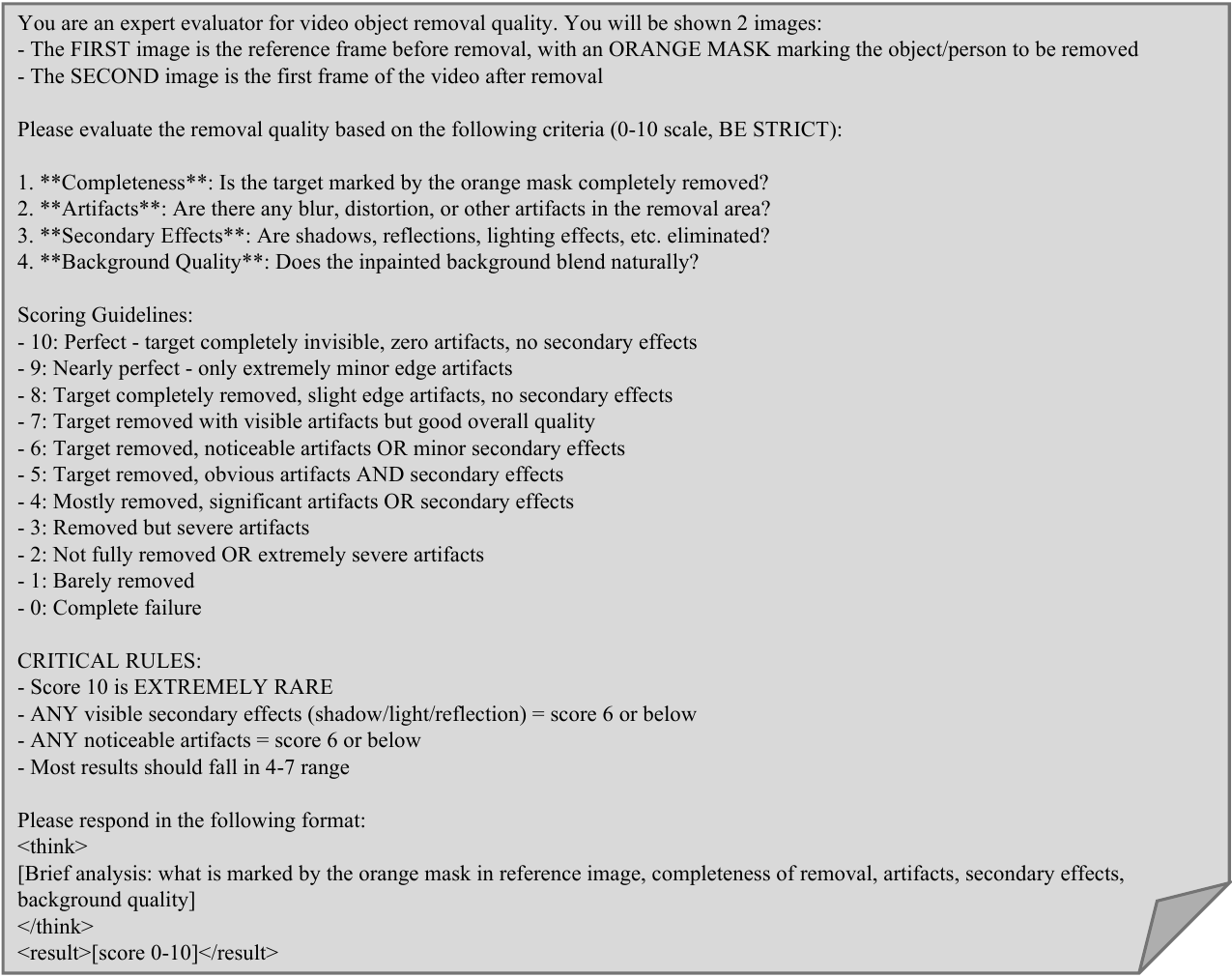}
\caption{\textbf{Prompt used for QScore evaluation.} The prompt guides Qwen-VL to assess removal completeness and visual artifacts.}
    \label{fig:prompt}
\end{figure*}

%% file: fig/fig_insert_more.tex
\begin{figure*}[]
    \centering
    \includegraphics[width=0.992\textwidth]{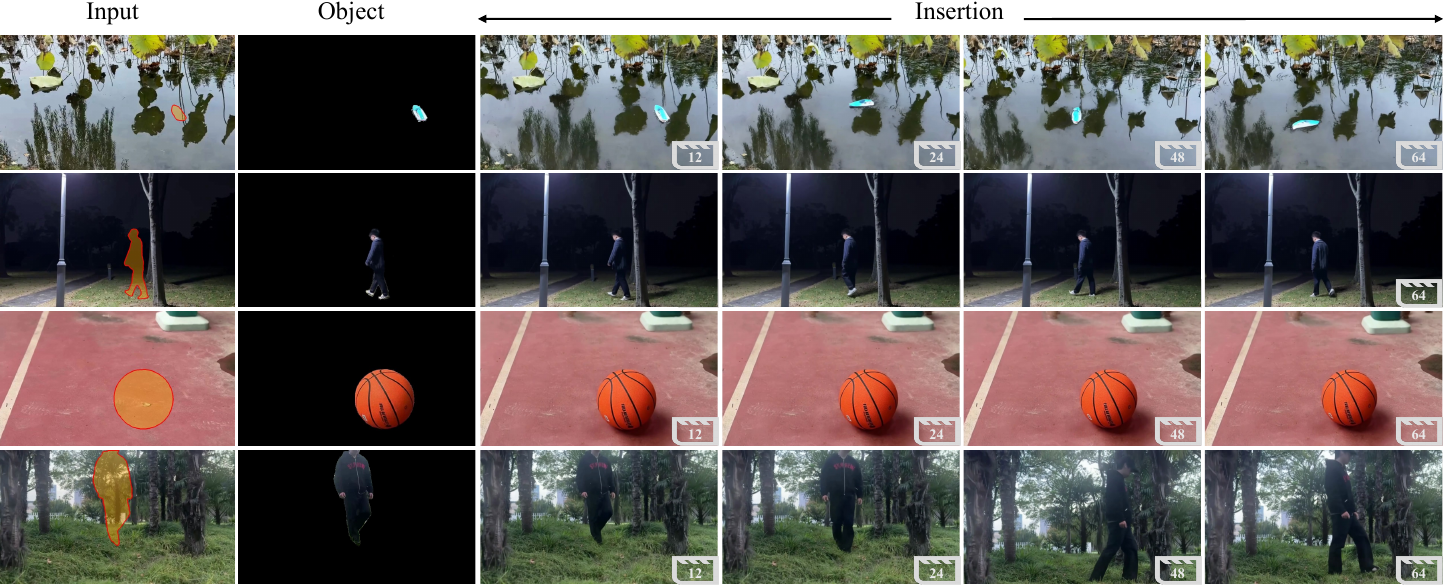}
    \caption{{More insertion results of EffectErase.}
    }
    \label{fig:insert_more}
\end{figure*}

%% file: fig/fig_com_more.tex
\begin{figure*}[]
    \centering
    \includegraphics[width=0.992\textwidth]{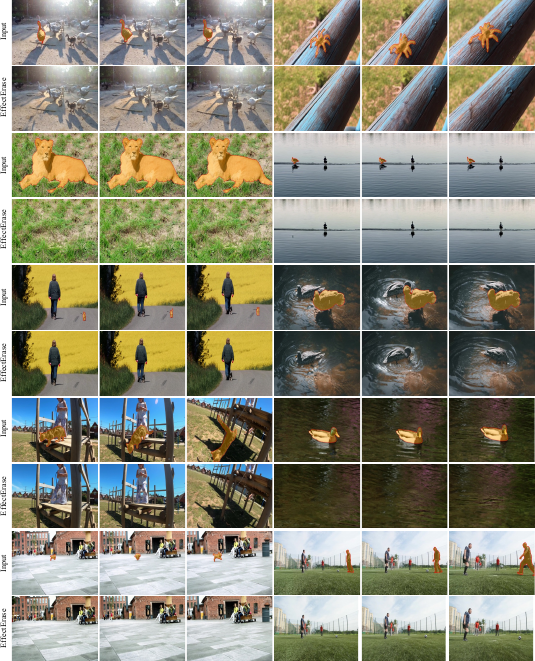}
    \caption{{More removal results of EffectErase.}
    }
    \label{fig:comp_more}
\end{figure*}

%% file: fig/fig_comp_vor.tex
\begin{figure*}[t]
    \centering
    \includegraphics[width=0.992\textwidth]{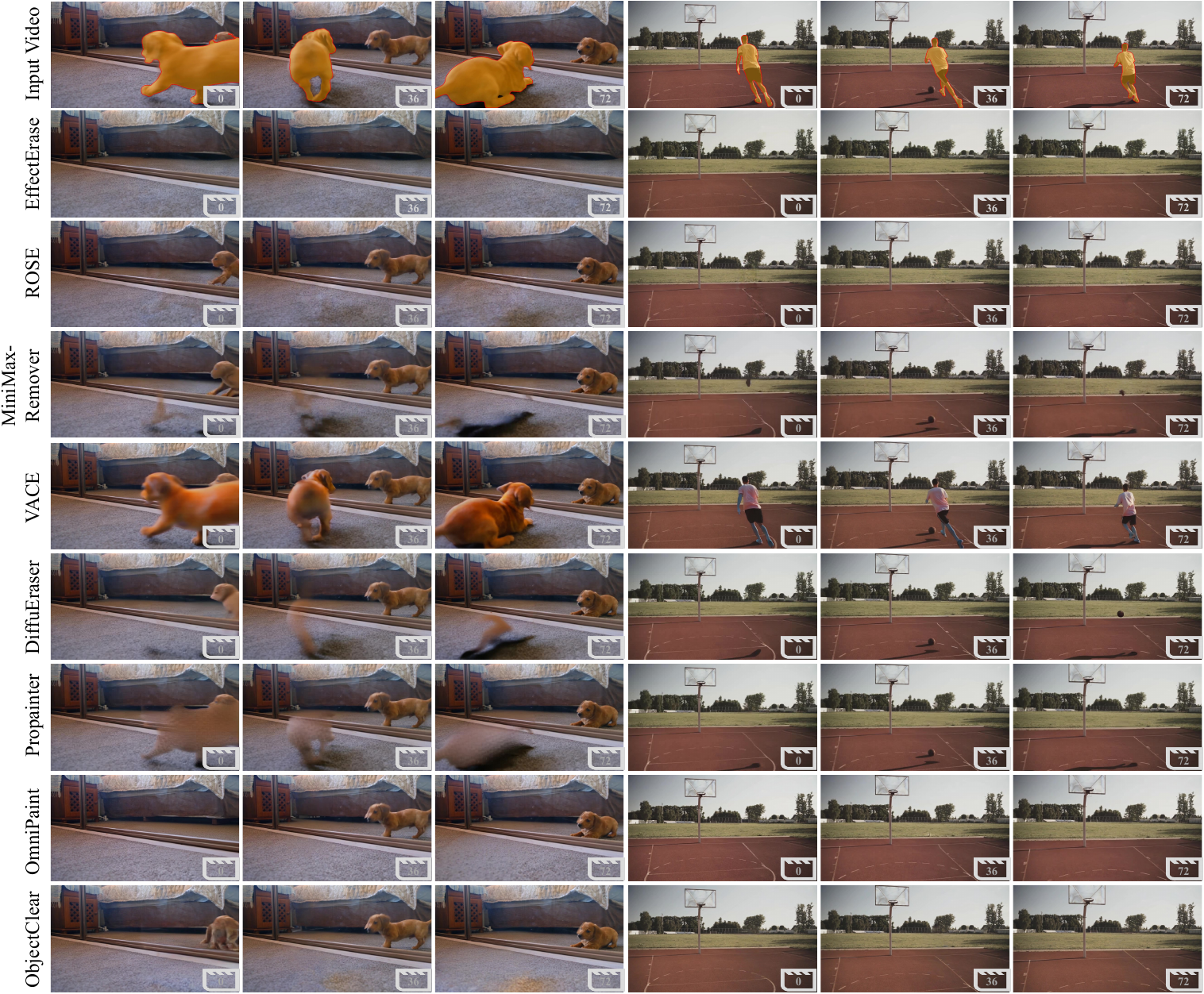}
    \caption{More comparison with state-of-the-art methods. }
    
    \label{fig:comp_vor}
\end{figure*}

%% file: main.bib
@String(CVPR= {IEEE Conf. Comput. Vis. Pattern Recog.})

@String(ICCV= {Int. Conf. Comput. Vis.})

@String(ECCV= {Eur. Conf. Comput. Vis.})

@String(NeurIPS= {Adv. Neural Inform. Process. Syst.})

@String(ICPR = {Int. Conf. Pattern Recog.})

@String(BMVC= {Brit. Mach. Vis. Conf.})

@String(TIP  = {IEEE Trans. Image Process.})

@String(ICLR = {Int. Conf. Learn. Represent.})

@String(AAAI = {AAAI})

@String(CVPRW= {IEEE Conf. Comput. Vis. Pattern Recog. Worksh.})

@String(CVPR  = {CVPR})

@String(ICCV  = {ICCV})

@String(ECCV  = {ECCV})

@String(NeurIPS  = {NeurIPS})

@String(ICPR  = {ICPR})

@String(BMVC  =	{BMVC})

@String(TIP   = {IEEE TIP})

@String(ICLR  = {ICLR})

@String(CVPRW= {CVPRW})

@String(ICML = {ICML})

@String(JMLR = {JMLR})

@article{brooks2024video,
  title={Video generation models as world simulators},
  author={Brooks, Tim and Peebles, Bill and Holmes, Connor and DePue, Will and Guo, Yufei and Jing, Li and Schnurr, David and Taylor, Joe and Luhman, Troy and Luhman, Eric and others},
  journal={OpenAI Blog},
  volume={1},
  number={8},
  xxxxpages={1},
  year={2024}
}

@article{yang2024cogvideox,
  title={Cogvideox: Text-to-video diffusion models with an expert transformer},
  author={Yang, Zhuoyi and Teng, Jiayan and Zheng, Wendi and Ding, Ming and Huang, Shiyu and Xu, Jiazheng and Yang, Yuanming and Hong, Wenyi and Zhang, Xiaohan and Feng, Guanyu and others},
  journal={arXiv preprint arXiv:2408.06072},
  year={2024}
}

@article{kong2024hunyuanvideo,
  title={Hunyuanvideo: A systematic framework for large video generative models},
  author={Kong, Weijie and Tian, Qi and Zhang, Zijian and Min, Rox and Dai, Zuozhuo and Zhou, Jin and Xiong, Jiangfeng and Li, Xin and Wu, Bo and Zhang, Jianwei and others},
  journal={arXiv preprint arXiv:2412.03603},
  year={2024}
}

@article{wan2025wan,
  title={Wan: Open and advanced large-scale video generative models},
  author={Wan, Team and Wang, Ang and Ai, Baole and Wen, Bin and Mao, Chaojie and Xie, Chen-Wei and Chen, Di and Yu, Feiwu and Zhao, Haiming and Yang, Jianxiao and others},
  journal={arXiv preprint arXiv:2503.20314},
  year={2025}
}

@inproceedings{wang2019video,
  title={Video inpainting by jointly learning temporal structure and spatial details},
  author={Wang, Chuan and Huang, Haibin and Han, Xiaoguang and Wang, Jue},
  booktitle=AAAI,
  xxxxpages={5232--5239},
  year={2019}
}

@inproceedings{chang2019free,
  title={Free-form video inpainting with 3d gated convolution and temporal patchgan},
  author={Chang, Ya-Liang and Liu, Zhe Yu and Lee, Kuan-Ying and Hsu, Winston},
  booktitle=CVPR,
  xxxxpages={9066--9075},
  year={2019}
}

@inproceedings{liu2021fuseformer,
  title={Fuseformer: Fusing fine-grained information in transformers for video inpainting},
  author={Liu, Rui and Deng, Hanming and Huang, Yangyi and Shi, Xiaoyu and Lu, Lewei and Sun, Wenxiu and Wang, Xiaogang and Dai, Jifeng and Li, Hongsheng},
  booktitle=CVPR,
  xxxxpages={14040--14049},
  year={2021}
}

@inproceedings{zhang2022flow,
  title={Flow-guided transformer for video inpainting},
  author={Zhang, Kaidong and Fu, Jingjing and Liu, Dong},
  booktitle=ECCV,
  xxxxpages={74--90},
  year={2022},
  xxxxorganization={Springer}
}

@inproceedings{zhou2023propainter,
  title={Propainter: Improving propagation and transformer for video inpainting},
  author={Zhou, Shangchen and Li, Chongyi and Chan, Kelvin CK and Loy, Chen Change},
  booktitle=CVPR,
  xxxxpages={10477--10486},
  year={2023}
}

@article{li2025diffueraser,
  title={Diffueraser: A diffusion model for video inpainting},
  author={Li, Xiaowen and Xue, Haolan and Ren, Peiran and Bo, Liefeng},
  journal={arXiv preprint arXiv:2501.10018},
  year={2025}
}

@article{bian2025videopainter,
  title={VideoPainter: Any-length Video Inpainting and Editing with Plug-and-Play Context Control},
  author={Bian, Yuxuan and Zhang, Zhaoyang and Ju, Xuan and Cao, Mingdeng and Xie, Liangbin and Shan, Ying and Xu, Qiang},
  journal={arXiv preprint arXiv:2503.05639},
  year={2025}
}

@article{vace,
    title = {VACE: All-in-One Video Creation and Editing},
    author = {Jiang, Zeyinzi and Han, Zhen and Mao, Chaojie and Zhang, Jingfeng and Pan, Yulin and Liu, Yu},
    journal = {arXiv preprint arXiv:2503.07598},
    year = {2025}
}

@inproceedings{Sagong_2022_BMVC,
author    = {Min-Cheol Sagong and Yoon-Jae Yeo and Seung-Won Jung and Sung-Jea Ko},
title     = {RORD: A Real-world Object Removal Dataset},
booktitle = BMVC,
xxxxpublisher = {{BMVA} Press},
year      = {2022},
url       = {https://bmvc2022.mpi-inf.mpg.de/0542.pdf}
}

@inproceedings{liu2024shadow,
  title={Shadow generation for composite image using diffusion model},
  author={Liu, Qingyang and You, Junqi and Wang, Jianting and Tao, Xinhao and Zhang, Bo and Niu, Li},
  booktitle=CVPR,
  xxxxpages={8121--8130},
  year={2024}
}

@article{MOSEv2,
  title={{MOSEv2}: A More Challenging Dataset for Video Object Segmentation in Complex Scenes},
  author={Ding, Henghui and Ying, Kaining and Liu, Chang and He, Shuting and Jiang, Xudong and Jiang, Yu-Gang and Torr, Philip HS and Bai, Song},
  journal={arXiv preprint arXiv:2508.05630},
  year={2025}
}

@inproceedings{MOSE,
  title={{MOSE}: A New Dataset for Video Object Segmentation in Complex Scenes},
  author={Ding, Henghui and Liu, Chang and He, Shuting and Jiang, Xudong and Torr, Philip HS and Bai, Song},
  booktitle=ICCV,
  year={2023}
}

@article{wei2025omnieraser,
  title={OmniEraser: Remove Objects and Their Effects in Images with Paired Video-Frame Data},
  author={Wei, Runpu and Yin, Zijin and Zhang, Shuo and Zhou, Lanxiang and Wang, Xueyi and Ban, Chao and Cao, Tianwei and Sun, Hao and He, Zhongjiang and Liang, Kongming and others},
  journal={arXiv preprint arXiv:2501.07397},
  year={2025}
}

@article{zhao2025objectclear,
  title={ObjectClear: Complete Object Removal via Object-Effect Attention},
  author={Zhao, Jixin and Zhou, Shangchen and Wang, Zhouxia and Yang, Peiqing and Loy, Chen Change},
  journal={arXiv preprint arXiv:2505.22636},
  year={2025}
}

@inproceedings{chang2019vornet,
  title={Vornet: Spatio-temporally consistent video inpainting for object removal},
  author={Chang, Ya-Liang and Yu Liu, Zhe and Hsu, Winston},
  booktitle=CVPRW,
  xxxxpages={0--0},
  year={2019}
}

@article{xu2018youtube,
  title={Youtube-vos: A large-scale video object segmentation benchmark},
  author={Xu, Ning and Yang, Linjie and Fan, Yuchen and Yue, Dingcheng and Liang, Yuchen and Yang, Jianchao and Huang, Thomas},
  journal={arXiv preprint arXiv:1809.03327},
  year={2018}
}

@article{gu2024coherent,
  title={Coherent Video Inpainting Using Optical Flow-Guided Efficient Diffusion},
  author={Gu, Bohai and Luo, Hao and Guo, Song and Dong, Peiran and Zhou, Qihua},
  journal={arXiv preprint arXiv:2412.00857},
  year={2024}
}

@inproceedings{winter2024objectdrop,
  title={Objectdrop: Bootstrapping counterfactuals for photorealistic object removal and insertion},
  author={Winter, Daniel and Cohen, Matan and Fruchter, Shlomi and Pritch, Yael and Rav-Acha, Alex and Hoshen, Yedid},
  booktitle=ECCV,
  xxxxpages={112--129},
  year={2024},
  xxxxorganization={Springer}
}

@inproceedings{zi2025cococo,
  title={Cococo: Improving text-guided video inpainting for better consistency, controllability and compatibility},
  author={Zi, Bojia and Zhao, Shihao and Qi, Xianbiao and Wang, Jianan and Shi, Yukai and Chen, Qianyu and Liang, Bin and Xiao, Rong and Wong, Kam-Fai and Zhang, Lei},
  booktitle=AAAI,
  xxxxpages={11067--11076},
  year={2025}
}

@inproceedings{jiang2025smarteraser,
  title={Smarteraser: Remove anything from images using masked-region guidance},
  author={Jiang, Longtao and Wang, Zhendong and Bao, Jianmin and Zhou, Wengang and Chen, Dongdong and Shi, Lei and Chen, Dong and Li, Houqiang},
  booktitle=CVPR,
  xxxxpages={24452--24462},
  year={2025}
}

@inproceedings{liu2025erase,
  title={Erase Diffusion: Empowering Object Removal Through Calibrating Diffusion Pathways},
  author={Liu, Yi and Zhou, Hao and Cui, Benlei and Shang, Wenxiang and Lin, Ran},
  booktitle=CVPR,
  xxxxpages={2418--2427},
  year={2025}
}

@inproceedings{yang2025generative,
  title={Generative Image Layer Decomposition with Visual Effects},
  author={Yang, Jinrui and Liu, Qing and Li, Yijun and Kim, Soo Ye and Pakhomov, Daniil and Ren, Mengwei and Zhang, Jianming and Lin, Zhe and Xie, Cihang and Zhou, Yuyin},
  booktitle=CVPR,
  xxxxpages={7643--7653},
  year={2025}
}

@article{yu2025omnipaint,
  title={Omnipaint: Mastering object-oriented editing via disentangled insertion-removal inpainting},
  author={Yu, Yongsheng and Zeng, Ziyun and Zheng, Haitian and Luo, Jiebo},
  journal={arXiv preprint arXiv:2503.08677},
  year={2025}
}

@inproceedings{li2025rorem,
  title={RORem: Training a Robust Object Remover with Human-in-the-Loop},
  author={Li, Ruibin and Yang, Tao and Guo, Song and Zhang, Lei},
  booktitle=CVPR,
  xxxxpages={14024--14035},
  year={2025}
}

@misc{zi2025minimaxremovertamingbadnoise,
      title={MiniMax-Remover: Taming Bad Noise Helps Video Object Removal}, 
      author={Bojia Zi and Weixuan Peng and Xianbiao Qi and Jianan Wang and Shihao Zhao and Rong Xiao and Kam-Fai Wong},
      year={2025},
      eprint={2505.24873},
      archivePrefix={arXiv},
      primaryClass={cs.CV},
      url={https://arxiv.org/abs/2505.24873}, 
}

@inproceedings{hore2010image,
  title={Image quality metrics: PSNR vs. SSIM},
  author={Hore, Alain and Ziou, Djemel},
  booktitle={ICPR},
  xxxxpages={2366--2369},
  year={2010},
  xxxxorganization={IEEE}
}

@article{wang2004image,
  title={Image quality assessment: from error visibility to structural similarity},
  author={Wang, Zhou and Bovik, Alan C and Sheikh, Hamid R and Simoncelli, Eero P},
  journal=TIP,
  volume={13},
  number={4},
  xxxxpages={600--612},
  year={2004},
  xxxxpublisher={IEEE}
}

@inproceedings{zhang2018unreasonable,
  title={The unreasonable effectiveness of deep features as a perceptual metric},
  author={Zhang, Richard and Isola, Phillip and Efros, Alexei A and Shechtman, Eli and Wang, Oliver},
  booktitle=CVPR,
  xxxxpages={586--595},
  year={2018}
}

@inproceedings{unterthiner2019fvd,
  title={FVD: A new metric for video generation},
  author={Unterthiner, Thomas and Van Steenkiste, Sjoerd and Kurach, Karol and Marinier, Rapha{\"e}l and Michalski, Marcin and Gelly, Sylvain},
  booktitle={ICLR Workshop},
  year={2019}
}

@inproceedings{hu2022lora,
  title={Lora: Low-rank adaptation of large language models},
  author={Hu, Edward J and Shen, Yelong and Wallis, Phillip and Allen-Zhu, Zeyuan and Li, Yuanzhi and Wang, Shean and Wang, Lu and Chen, Weizhu and others},
  booktitle=ICLR,
  year={2022}
}

@inproceedings{ravi2024sam2,
  title={SAM 2: Segment Anything in Images and Videos},
  author={Ravi, Nikhila and Gabeur, Valentin and Hu, Yuan-Ting and Hu, Ronghang and Ryali, Chaitanya and Ma, Tengyu and Khedr, Haitham and R{\"a}dle, Roman and Rolland, Chloe and Gustafson, Laura and Mintun, Eric and Pan, Junting and Alwala, Kalyan Vasudev and Carion, Nicolas and Wu, Chao-Yuan and Girshick, Ross and Doll{\'a}r, Piotr and Feichtenhofer, Christoph},
  booktitle=ICLR,
  year={2025}
}

@article{wan2025,
      title={Wan: Open and Advanced Large-Scale Video Generative Models}, 
      author={Team Wan and Ang Wang and Baole Ai and Bin Wen and et al},
      journal = {arXiv preprint arXiv:2503.20314},
      year={2025}
}

@article{hendrycks2016gaussian,
  title={Gaussian Error Linear Units (Gelus)},
  author={Hendrycks, D},
  journal={arXiv preprint arXiv:1606.08415},
  year={2016}
}

@inproceedings{Peebles2022DiT,
  title={Scalable diffusion models with transformers},
  author={Peebles, William and Xie, Saining},
  booktitle=ICCV,
  pages={4195--4205},
  year={2023}
}

@inproceedings{glorot2010understanding,
  author       = {Xavier Glorot and
                  Yoshua Bengio},
  editor       = {Yee Whye Teh and
                  D. Mike Titterington},
  title        = {Understanding the difficulty of training deep feedforward neural networks},
  booktitle    = {AISTATS},
  series       = {{JMLR} Proceedings},
  volume       = {9},
  pages        = {249--256},
  year         = {2010},
}

@article{ba2016layer,
  title={Layer normalization},
  author={Ba, Jimmy Lei and Kiros, Jamie Ryan and Hinton, Geoffrey E},
  journal={arXiv preprint arXiv:1607.06450},
  year={2016}
}

@inproceedings{adamw,
  author       = {Ilya Loshchilov and
                  Frank Hutter},
  title        = {Decoupled Weight Decay Regularization},
  booktitle    = ICLR,
  year         = {2019},
}

@inproceedings{he2015delving,
  title={Delving deep into rectifiers: Surpassing human-level performance on imagenet classification},
  author={He, Kaiming and Zhang, Xiangyu and Ren, Shaoqing and Sun, Jian},
  booktitle=ICCV,
  pages={1026--1034},
  year={2015}
}

@article{kingma2013auto,
  title={Auto-encoding variational bayes},
  author={Kingma, Diederik P and Welling, Max},
  journal={arXiv preprint arXiv:1312.6114},
  year={2013}
}

@article{raffel2020exploring,
  title={Exploring the limits of transfer learning with a unified text-to-text transformer},
  author={Raffel, Colin and Shazeer, Noam and Roberts, Adam and Lee, Katherine and Narang, Sharan and Matena, Michael and Zhou, Yanqi and Li, Wei and Liu, Peter J},
  journal=JMLR,
  volume={21},
  number={140},
  xxxxpages={1--67},
  year={2020}
}

@inproceedings{radford2021learning,
  title={Learning transferable visual models from natural language supervision},
  author={Radford, Alec and Kim, Jong Wook and Hallacy, Chris and Ramesh, Aditya and Goh, Gabriel and Agarwal, Sandhini and Sastry, Girish and Askell, Amanda and Mishkin, Pamela and Clark, Jack and others},
  booktitle=ICML,
  xxxxpages={8748--8763},
  year={2021},
  xxxxorganization={PmLR}
}

@inproceedings{esser2024scaling,
  title={Scaling rectified flow transformers for high-resolution image synthesis},
  author={Esser, Patrick and Kulal, Sumith and Blattmann, Andreas and Entezari, Rahim and M{\"u}ller, Jonas and Saini, Harry and Levi, Yam and Lorenz, Dominik and Sauer, Axel and Boesel, Frederic and others},
  booktitle=ICML,
  year={2024}
}

@inproceedings{vaswani2017attention,
  title={Attention is all you need},
  author={Vaswani, Ashish and Shazeer, Noam and Parmar, Niki and Uszkoreit, Jakob and Jones, Llion and Gomez, Aidan N and Kaiser, {\L}ukasz and Polosukhin, Illia},
  booktitle=NeurIPS,
  year={2017}
}

@inproceedings{
miao2025rose,
title={{ROSE}: Remove Objects with Side Effects in Videos},
author={Chenxuan Miao and Yutong Feng and Jianshu Zeng and Zixiang Gao and Liu Hantang and Yunfeng Yan and Donglian Qi and Xi Chen and Bin Wang and Hengshuang Zhao},
booktitle=NeurIPS,
year={2025},
}

@article{bai2025qwen2,
  title={Qwen2. 5-vl technical report},
  author={Bai, Shuai and Chen, Keqin and Liu, Xuejing and Wang, Jialin and Ge, Wenbin and Song, Sibo and Dang, Kai and Wang, Peng and Wang, Shijie and Tang, Jun and others},
  journal={arXiv preprint arXiv:2502.13923},
  year={2025}
}
